\pdfoutput=1
\documentclass{article}

\usepackage{microtype}
\usepackage{graphicx}
\usepackage{subfigure}
\usepackage{booktabs} % for professional tables

\usepackage{hyperref}

\usepackage[accepted]{icml2023}

\usepackage{amsmath}
\usepackage{amssymb}
\usepackage{mathtools}
\usepackage{amsthm}

\usepackage[capitalize,noabbrev]{cleveref}

\theoremstyle{plain}

\theoremstyle{definition}

\theoremstyle{remark}

\usepackage[textsize=tiny]{todonotes}

\usepackage{graphicx}
\usepackage{colortbl}
\usepackage{pifont}
\usepackage{bbding}   
\usepackage{fontawesome}  
\usepackage[algo2e,ruled,noend]{algorithm2e}
\usepackage{soul}
\usepackage{multirow}
\usepackage{makecell}
\newcommand{\cmark}{\ding{51}}
\newcommand{\xmark}{\ding{55}}

\newcommand{\ie}{\textit{i}.\textit{e}.}
\newcommand{\eg}{\textit{e}.\textit{g}.}
\usepackage{caption}
\definecolor{ForestGreen}{RGB}{34,139,34}
\definecolor{brickred}{rgb}{0.8, 0.25, 0.33}
\usepackage{rotating}  

\usepackage{bm}
\usepackage{color}
\definecolor{highlight}{RGB}{200,210,255}
\usepackage{hyperref}

\icmltitlerunning{UPop: Unified and Progressive Pruning for Compressing Vision-Language Transformers}

\begin{document}

\twocolumn[
\icmltitle{UPop: Unified and Progressive Pruning for \\
Compressing Vision-Language Transformers}



\icmlsetsymbol{correspondence}{$\dagger$}

\begin{icmlauthorlist}
\icmlauthor{Dachuan Shi}{thu,shlab}
\icmlauthor{Chaofan Tao}{hku}
\icmlauthor{Ying Jin}{cuhk}
\icmlauthor{Zhendong Yang}{thu}
\icmlauthor{Chun Yuan}{thu,correspondence}
\icmlauthor{Jiaqi Wang}{shlab,correspondence}

\end{icmlauthorlist}

\icmlaffiliation{thu}{Tsinghua University}
\icmlaffiliation{shlab}{Shanghai AI Laboratory}
\icmlaffiliation{hku}{The University of Hong Kong}
\icmlaffiliation{cuhk}{The Chinese University of Hong Kong}

\icmlcorrespondingauthor{Chun Yuan}{yuanc@sz.tsinghua.edu.cn}
\icmlcorrespondingauthor{Jiaqi Wang}{wjqdev@gmail.com}

\icmlkeywords{Multimodal Model, Model Compression, Vision-Language Transformers}

\vskip 0.3in
]



\printAffiliationsAndNotice{}  % leave blank if no need to mention equal contribution

%%%%%%%%%%%%%%%%%%%%%%%%%%%%%%%%%%%%%%%%%%%%%%%%%%%%%%%%%%%%%%%%%%%%%%%%%%%%%%%
%%%%%%%%%%%%%%%%%%%%%%%%%%%%%%%%%%%%%%%%%%%%%%%%%%%%%%%%%%%%%%%%%%%%%%%%%%%%%%%
% MAIN TEXT
%%%%%%%%%%%%%%%%%%%%%%%%%%%%%%%%%%%%%%%%%%%%%%%%%%%%%%%%%%%%%%%%%%%%%%%%%%%%%%%
%%%%%%%%%%%%%%%%%%%%%%%%%%%%%%%%%%%%%%%%%%%%%%%%%%%%%%%%%%%%%%%%%%%%%%%%%%%%%%%

\begin{abstract}
Real-world data contains a vast amount of multimodal information, among which vision and language are the two most representative modalities. Moreover, increasingly heavier models, \textit{e}.\textit{g}., Transformers, have attracted the attention of researchers to model compression. However, how to compress multimodal models, especially vison-language Transformers, is still under-explored. This paper proposes the \textbf{U}nified and \textbf{P}r\textbf{o}gressive \textbf{P}runing (\textbf{\emph{UPop}}) as a universal vison-language Transformer compression framework, which incorporates 1) unifiedly searching multimodal subnets in a continuous optimization space from the original model, which enables automatic assignment of pruning ratios among compressible modalities and structures; 2) progressively searching and retraining the subnet, which maintains convergence between the search and retrain to attain higher compression ratios. Experiments on various tasks, datasets, and model architectures demonstrate the effectiveness and versatility of the proposed UPop framework. The code is available at \href{https://github.com/sdc17/UPop}{https://github.com/sdc17/UPop}.

\end{abstract}

\section{Introduction}
\label{INTRODUCTION}

The number of parameters and FLOPs of deep learning models \cite{devlin2018bert, shoeybi2019megatron, brown2020language, shao2021intern, smith2022using} have proliferated in recent years, which makes compression exceedingly critical for deploying the increasingly heavier models on edge devices. There are lots of approaches for model compression, such as weight sharing \cite{lan2019albert}, low-rank factorization \cite{yu2017compressing}, quantization \cite{tao2022compression}, parameter bootstrapping \cite{chen2022litevl}, knowledge distillation \cite{yang2022masked}, 
and pruning \cite{han2015learning}. As the paradigm this paper focuses on, pruning approaches not only benefit from inheriting well-optimized parameters of the original model but also provide flexible design space for various architectures.

Recently, pruning approaches dedicated to the Transformers \cite{vaswani2017attention} have attracted much attention. According to the pruned components, these approaches can be summarized into two categories. 1) Token Pruning: By eliminating the number of input tokens, these approaches \cite{goyal2020power, rao2021dynamicvit} can reduce the FLOPs of models. 2) Model Pruning. By reducing the model size, these approaches \cite{chen2021chasing, su2022vitas} can reduce both the parameters and FLOPs of models. This paper focuses on model compression so that the parameters and FLOPs of models can be reduced simultaneously.

\begin{table}[t]
  \setlength{\tabcolsep}{1.5pt}
  \captionsetup{font={small}}
  \setlength{\abovecaptionskip}{0.1cm}
  \renewcommand{\arraystretch}{1.3}
  \scriptsize
  \centering  
  \label{table exp overview}
  \caption{Overview of experimental results at 2$\times$ compression. The proposed UPop framework is efficient and effective on various tasks, datasets, and architectures. \textbf{Bold} indicates the best post-compression performance. Mask-based Pruning is extended from the SOTA pruning method ViT-Slimming \cite{chavan2022vision}. }
  \begin{tabular}{l l l l l @{\hspace{0.7 \tabcolsep}} l @{\hspace{0.7 \tabcolsep}} l}
    \toprule
    \multirow{2}{*}{Method} & \multirow{2}{*}{\makecell{Visual \\ Reason}} & \multirow{2}{*}{\makecell{Image \\ Caption}} & \multirow{2}{*}{\makecell{Visual \\ QA}} & \multicolumn{2}{c}{Retrieval\quad\,} & \multirow{2}{*}{\makecell{Image Cla-\\ ssification}}\\
    % \cmidrule{5-6}
    & & & & COCO & Flickr & \\
    \midrule
    Original Model & 83.1 & 23.8 & 77.5 & 81.9 & 96.8 & 79.9\\
    % \midrule
    \rowcolor{gray!10} Mask-based Pruning & 76.4$_{\color{red}\downarrow 6.7}$ & 21.0$_{\color{red}\downarrow 2.8}$ & 71.6$_{\color{red}\downarrow 5.9}$ & 61.7$_{\color{red}\downarrow 20}$ & 78.9$_{\color{red}\downarrow 18}$ & 77.9$_{\color{red}\downarrow 2.0}$ \\
    % \midrule
    \rowcolor{highlight!18}  UPop (Ours) & \textbf{81.1}$_{\color{red}\downarrow 2.0}$ & \textbf{23.3}$_{\color{red}\downarrow 0.5}$ & \textbf{76.3}$_{\color{red}\downarrow 1.2}$ & \textbf{77.4}$_{\color{red}\downarrow 4.5}$ & \textbf{94.0}$_{\color{red}\downarrow 2.8}$ & \textbf{78.9}$_{\color{red}\downarrow 1.0}$ \\
    \bottomrule
  \end{tabular}
  \label{performance comparison}
  \vspace{-20pt}
\end{table}

\begin{figure*}[t]
    \centering
    \captionsetup{font={small}}
    \includegraphics[width=1.0\linewidth]{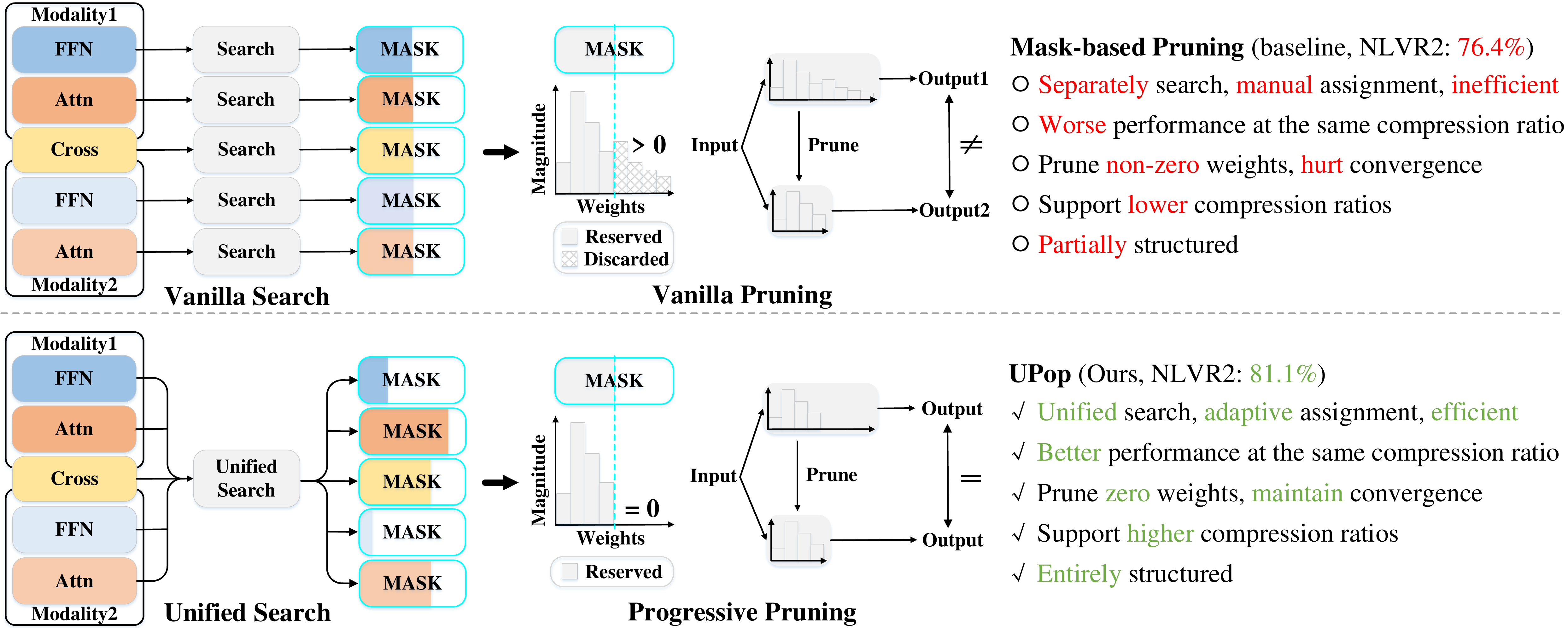}
    \vspace{-12pt}
    \caption{Comparison between the Mask-based Pruning (\eg, extending ViT-Slimming \cite{chavan2022vision} to the multimodal scenario) and our UPop framework. Mask-based Pruning manually assigns each compressible component with a predefined compression ratio, which is inefficient and sub-optimal. Moreover, the vanilla pruning paradigm fails when it comes to higher compression ratios. UPop enables adaptively assigning the pruning ratio to each compressible component, which achieves significant performance improvements at the same overall compression ratio. Moreover, the progressive pruning paradigm eliminates the weight gap between the searched model and the pruned subnet to be retrained, therefore gaining better convergence and performance, especially at high compression ratios.}
    \label{figure comparison}
    \vspace{-14pt}
\end{figure*}

In real applications, there are lots of multimodal tasks that have been extensively studied, including but not limited to Visual Question Answer \citep{antol2015vqa}, Image Caption \citep{lin2014microsoft}, and Image-Text Retrieval \citep{https://doi.org/10.48550/arxiv.1509.04942}. To tackle these multimodal tasks, various multimodal models \cite{kiros2014unifying, karpathy2014deep, antol2015vqa, vinyals2015show, yang2016stacked, huang2017instance} have been proposed accordingly. Furthermore, as Transformer \cite{vaswani2017attention} has been more and more popular among deep models, transformer-based models \cite{tan2019lxmert, lu2019vilbert, zhou2020unified, li2020oscar, kim2021vilt, jia2021scaling, yu2022coca, wang2022image} have also dominated the recent studies of multimodal models. For example, CLIP \cite{radford2021learning} and BLIP \cite{li2022blip} are some of the most representative ones. Benefiting from massive image-text pairs as pre-training datasets, they can learn joint representations of multiple modalities and can be further used to fine-tune on downstream tasks. 

Although compression on unimodal tasks has been widely investigated, how to compress multimodal models, especially vision-language Transformers, is still relatively under-explored. In this paper, we propose a novel multimodal pruning framework, Unified and Progressive Pruning (UPop).

\begin{table*}[t]
\begin{small}
  \centering
  \captionsetup{font={small}}
  \caption{Here we list the notations table. In the later part of the article, superscript $^{\{v,l,c\}}$\ indicates notations for vision, language, and cross-modality, respectively, subscript $_{\{a,m\}}$\ indicates notations for Attention and MLP structure, respectively.}
  \label{Notation}
  \vspace{-6pt}
  \begin{tabular*}{\linewidth}{l @{\hspace{4\tabcolsep}} l @{\hspace{7\tabcolsep}} l @{\hspace{4\tabcolsep}} l}
    \toprule
    \bf{NOTATION}  & \bf{DESCRIPTION}  & \bf{NOTATION}  & \bf{DESCRIPTION} \\
    \midrule
    $L$ & Number of layers & $H$ & Number of heads \\
    $N$ & Number of patches / Sequence length  & $D$ & Embedding size \\
    $d$ & Embedding size of each head & $p$ & Total compression ratio \\
    $\bm{\theta}$ & Parameters of the original model & $\bm{\zeta}$ & Parameters of the trainable mask \\
    $w$ & Regularization coefficient in searching & $\mathcal{F}_p$ & $p\%$ compressed model $\mathcal{F}_p(x|\bm{\theta}, \bm{\zeta})$ \\
    $\alpha,\ \beta$ & Learning rate during$\{$search, retrain$\}$ & $T_{\{s,r\}}$ & Iterations in $\{$search, retrain$\}$ phase \\
    \bottomrule
    \vspace{-16pt}
  \end{tabular*}
 \end{small}
\end{table*}

A straightforward design of multimodal pruning is to compress each modality separately via the unimodal pruning approach. However, there exist two main challenges. One of the challenges is that we have to manually explore suitable compression ratios for different components in different modalities, which is inefficient, especially when the model has multiple types of modules (these modules may comprise Self-Attentions, Cross-Attentions, and Feed-Forward Nets in both vision and language branches for typical vision-language Transformers). Moreover, when given a total compression ratio, the optimal compression ratio for different modalities and modules may vary, and therefore manual assignment is most likely sub-optimal. To overcome this shortcoming, we propose to unifiedly search on different modalities and different structures, which enables our approach to adaptively assign appropriate compression ratios among all compressible components given a total compression ratio. Comparison is illustrated in Figure \ref{figure comparison} (Vanilla Search vs. Unified Search).

The second challenge is that the traditional two-stage pruning paradigm (\ie, retraining after searching) fails when the compression ratio is high. After the search stage, unimportant neurons are going to be removed. However, many of them have non-zero weights, and suddenly binarizing them to zero after searching harms the convergence of the pruned subnet. In other words, the significant gap of parameter weights between the searched model (\ie, model after the searching stage) and the pruned subnet to be retrained cause it is hard to converge and severely degrades the final performance. Consequently, we propose an improved pruning paradigm that conducts searching and retraining progressively and simultaneously, which ensures weights of removed neurons will progressively converge to zero before the end of the search stage, and therefore effectively eliminate the gap mentioned above. Comparison is illustrated in Figure \ref{figure comparison} (Vanilla Pruning vs. Progressive Pruning). 

Our main contributions can be summarized as
\vspace{-7pt}
\begin{itemize}
    \item In this paper, we propose a novel and universal multimodal pruning framework UPop for compressing vision-language Transformers. 
    \item The proposed \textit{Unified Pruning} enables adaptive compression ratio assignment among all compressible components. \textit{Progressive Pruning} proposes an improved pruning paradigm that gains better convergence and supports higher compression ratios.
    \item As a structured pruning framework, UPop's effectiveness and versatility are validated on various multimodal tasks, datasets, and model architectures (\eg, dual-stream CLIP \cite{radford2021learning} and mixed-stream BLIP \cite{li2022blip}), and also evaluated on unimodal tasks (\eg, image classification and segmentation). 
\end{itemize}

\section{Related Works}
\label{RELATED WORK}

\textbf{Vision-Language Transformer} Recently, significant progress in vision-language tasks has been achieved by various Vision-Lanauge Transformers \cite{radford2021learning, yu2022coca, wang2022image}, among which BLIP \cite{li2022blip} is one of the most representative models. BLIP is a pure transformed-based multimodal model, which employs a Bert \cite{devlin2018bert} and a ViT \cite{dosovitskiy2020image} as text encoder and image encoder, respectively. To allow multimodal interaction, BLIP injects vision information from the image encoder into the text encoder by inserting an additional cross-attention layer after the self-attention layer of each transformer block in the text encoder. 

\textbf{Transformer Pruning} There are several works exploring Transformers pruning on unimodal tasks. For example, structured pruning that removes layers \cite{fan2019reducing}, heads \cite{michel2019sixteen}, or channels \cite{zhu2021vision}, unstructured pruning \cite{yang2021nvit, chen2021chasing, sanh2020movement, chen2020lottery, liu2022win} that removes individual weights, and the intersection of structured and unstructured pruning such as ViT-Slimming \cite{chavan2022vision} that removes a different number of individual weights for different heads. UPop is a structured pruning approach whose minimum granularity is an entire row or column in the weights of model parameters. 

\textbf{Multimodal Transformer Compression} A few works have investigated the compression of multimodal Transformers. For example, MiniVLM \cite{wang2020minivlm} suggest an efficient feature extractor and a compact BERT \cite{devlin2018bert} as basic components of visual-language models. DistillVLM \cite{fang2021compressing} proposes that knowledge distillation can be used to mimic attention distributions from large vision-language models. The prior work \cite{gan2022playing} directly applies the unimodal pruning method \cite{han2015deep} on the multimodal scenario to verify whether the lottery tickets hypothesis also exists on multimodal models. The difference between UPop and \cite{gan2022playing} are 3 aspects: 1) \textit{Unified Pruning} enables adaptively instead of manually assigning the appropriate pruning ratio to each compressible component, 2) \textit{Progressive Pruning} gains better convergence and performance at high compression ratios. 3) UPop is structured and relatively easier to deploy, while \cite{gan2022playing} is unstructured and harder to deploy.

\textbf{Supplementary} Due to the space constraint, we provide more related works about global pruning, iterative pruning, and parameter-efficient tuning in Appendix \ref{more related work}.

\begin{figure*}[t]
    \centering
    \captionsetup{font={small}}
    \includegraphics[width=1.0\linewidth]{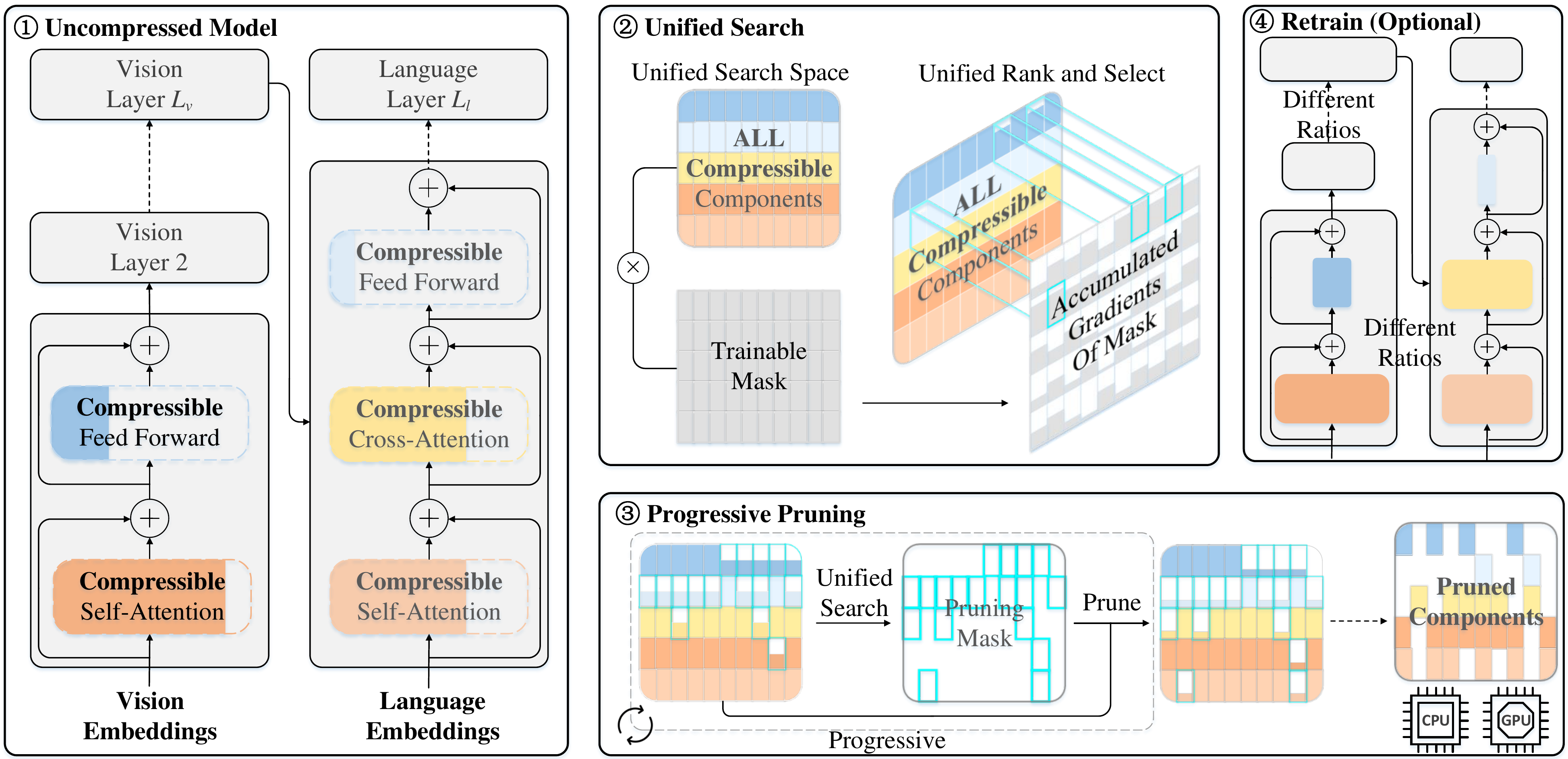}
    \vspace{-16pt}
    \caption{Diagram of \textit{Unified and Progressive Pruning} (UPop) framework. (1) Trainable masks are initialized to ones and inserted into Self-Attention, Cross-Attention, and MLP (Feed Forward Network) in each modality. (2) Combine all compressible components and trainable masks as a unified search space. Then, the current pruning mask is generated based on unified ranking and selecting the importance metric (\ie, accumulated gradients of the trainable masks). (3) Repeat the cycle consisting of unified search and progressive pruning until the target total compression ratio is reached. (4) Pruned subnet can be further fine-tuned to achieve better performance.}
    \label{figure UPop}
    \vspace{-10pt}
\end{figure*}

\section{Methodology}
\label{METHODOLOGY}

We propose \textit{Unified and Progressive Pruning} as illustrated in Figure \ref{figure UPop}. Necessary notations are listed in Table \ref{Notation}. We start by revisiting \textit{Mask-based Pruning} and straightforwardly extend it to the multimodal scenario.

\subsection{Mask-based Pruning}
\label{Mask-based Pruning}
Extended \textit{Mask-based Pruning} compresses vision and language Transformers separately via unimodal \textit{Mask-based Pruning}, consisting of a search phase and a retraining phase. Detailed implementation refers to Algorithm \ref{algorithm mask-based} in Appendix.

\textbf{Search} Take searching on Self-Attentions of Vision Transformer as an example. Denote the input of Self-Attention in the $l^{th}$ layer as $\bm{a}_{l} \in \mathbb{R}^{N\times D}$, and every head $h$ in the Self-Attention will transform $\bm{a}_{l}$ into query $\bm{q}_{l,h} \in \mathbb{R}^{N \times d}$, key $\bm{k}_{l,h} \in \mathbb{R}^{N \times d}$, and value $\bm{v}_{l,h} \in \mathbb{R}^{N \times d}$. The trainable mask $\bm{\zeta}_{a}^{v} \in \mathbb{R}^{L \times 1 \times d}$ will be initialized to ones and inserted into Self-Attentions of each layer.\footnote{More fine-grained mask shape $\mathbb{R}^{L \times H \times d}$ will result in pruned heads within a layer has different dimensions, and thus matrix computation of attention map becomes unfeasible on regular devices.} Then attention map of each head can be derived from
\begin{equation}
\setlength{\abovedisplayskip}{4pt}
\setlength{\belowdisplayskip}{-2pt}
\small
	\bm{A}_{l,h} = \text{Softmax}\left((\bm{q}_{l,h} \odot \bm{\zeta}_{a,l}^{v}) \times (\bm{k}_{l,h} \odot \bm{\zeta}_{a,l}^{v})^{\top}/ \sqrt{d}\right).
\end{equation}

\noindent The output of each head $h$ can be derived from
\begin{equation}
\setlength{\abovedisplayskip}{3pt}
\setlength{\belowdisplayskip}{-4pt}
\small
\bm{O}_{l,h} = \bm{A}_{l,h} \times (\bm{v}_{l,h} \odot \bm{\zeta}_{a,l}^{v}) \in \mathbb{R}^{N \times d}.
\end{equation}

\noindent Search on other structures (\eg, Cross-Attentions, FFNs) and modalities (\eg, vision, language) can be conducted similarly. Besides, the $\ell_{1}$-norm of masks $\bm{\zeta}$ are added as additional loss items to drive the magnitude of masks smaller:
\begin{equation}
\setlength{\abovedisplayskip}{2pt}
\setlength{\belowdisplayskip}{2pt}
\small
 \mathcal{L} = \mathcal{L_{O}} + w_a\sum\nolimits_{\bm{\zeta}_{i} \in \bm{\zeta}_a} \lVert \bm{\zeta}_{i} \rVert_{1} + w_m\sum\nolimits_{\bm{\zeta}_{i} \in \bm{\zeta}_m} \lVert \bm{\zeta}_{i} \rVert_{1}
\end{equation}
where $\mathcal{L_{O}}$ is the original loss to learn a multimodal model, and $w_a$ and $w_m$ are coefficients to balance the magnitude of loss items. It means that the model parameters $\bm{\theta}$ and trainable masks $\bm{\zeta}$ are optimized jointly in the search phase.

\textbf{Retraining} After the search, the subnet can be pruned from the searched model based on mask $\bm{\zeta}$. The magnitude of the mask is used as the metric to evaluate the importance of corresponding neurons. Neurons with the smallest magnitude of $p\%$ in the mask are removed (\ie, binarized as zero during retraining) from the searched model. The obtained subnet is retrained to get the final compressed model.

The major weakness of \textit{Mask-based Pruning} is two-fold: 1) the mask $\bm{\zeta}_i \in \bm{\zeta}$ on each module is assigned with a compression ratio manually, which is inefficient and sub-optimal, especially when the modules are usually various in a multimodal model; 2) for those neurons to be removed after search, their corresponding magnitude in the searched mask is not guaranteed to be zero. There are a lot of non-zero neurons with relatively small mask magnitudes, and suddenly binarizing them to zero after search harms the convergence of the pruned subnet. We tackle the aforementioned issues with \textit{Unified Pruning} and \textit{Progressive Pruning}, respectively.

\begin{figure*}[t]
\centering
\begin{minipage}[b]{\linewidth}
\begin{algorithm}[H]
    \small
    \caption{\small UPop: Unified and Progressive Pruning}
    \label{algorithm Upop}
    \setcounter{AlgoLine}{0}
    \LinesNumbered
    
    \KwIn{Original model $\mathcal{F}$, parameters of the original model $\bm{\theta}$, parameters of the trainable mask $\bm{\zeta}$, total compression ratio $p$, iterations in the search stage $T_{s}$ and retrain stage $T_{r}$, learning rate for the search stage $\alpha$ and retrain stage $\beta$ }

    \KwOut{Model $\mathcal{F^{\star}}$ after the search and retrain}
    
    \For{$t\gets0$ \KwTo $T_{r} - 1$} {
    
    \eIf{$t < T_{s}$}{

        \textcolor{gray}{\texttt{\# Calculate the loss $\mathcal{L}$, and normally update $\bm{\theta}$ with the original optimizer}}
        
        $\mathcal{L} \leftarrow \mathcal{L_{O}} + w_{a} \sum_{\bm{\zeta}_i \in \bm{\zeta}_a} \lVert \bm{\zeta}_i \rVert_{1} + w_{m} \sum_{\bm{\zeta}_i \in \bm{\zeta}_m} \lVert \bm{\zeta}_i \rVert_{1}$, \quad
        $ \bm{\theta}^{(t+1)} \leftarrow \bm{\theta}^{(t)} - \alpha \frac{1}{n} \sum_{i=1}^{n} \nabla_{\bm{\theta}} \mathcal{L}(\bm{\theta}^{(t)}, \bm{\zeta}^{(t)})$ 

        \textcolor{gray}{\texttt{\# Calculate the gradient of loss $\mathcal{L}$ with respect to the trainable mask $\bm{\zeta}$}}
        
        $\bm{G}^{(t)} \leftarrow \frac{1}{n} \sum_{i=1}^{n} \nabla_{\bm{\zeta}} \mathcal{L}(\bm{\theta}^{(t)}, \bm{\zeta}^{(t)}) $

        \textcolor{gray}{\texttt{\# Conduct z-score standardization to make the current $\bm{G}_a$ and $\bm{G}_m$ comparable}}
        
        $\bm{G}_a^{(t)} \leftarrow (\bm{G}_a^{(t)}-\mathbb{E}[\bm{G}_a^{(t)}]) / (\mathbb{E}[[\bm{G}_a^{(t)} - \mathbb{E}[\bm{G}_a^{(t)}]]^2])^{\frac{1}{2}} $,\quad 
        $\bm{G}_m^{(t)} \leftarrow (\bm{G}_m^{(t)}-\mathbb{E}[\bm{G}_m^{(t)}]) / (\mathbb{E}[[\bm{G}_m^{(t)} - \mathbb{E}[\bm{G}_m^{(t)}]]^2])^{\frac{1}{2}} $

        \textcolor{gray}{\texttt{\# Generate pruning mask $\bm{M}_t$ by ranking and selecting on accumulated gradient $\sum_{i=0}^{t} \bm{G}^{(i)}$}}
        
        $p_t \leftarrow p(\frac{1}{2}(1-\small\text{cos}(\frac{\pi t}{T_s-1})))^{\frac{1}{2}}$, \quad
        $\bm{M}^t \leftarrow {\tt\small TopKMask}(\sum_{i=0}^{t} \bm{G}^{(i)}, \ p_t \cdot {\tt\small Size} (\bm{\zeta}))$
        
        \textcolor{gray}{\texttt{\# Progressively compress $\bm{\zeta}_t$ based on $M_t$ and accordingly progressively compress $\mathcal{F}$}}
        
        $\bm{\zeta}^{(t+1)} \leftarrow (1 - \bm{M}^t) + (1 - \frac{p_t}{p})\bm{M}^t $, \quad
        $\mathcal{F}_{p_{t+1}} \leftarrow \mathcal{F}_{p_t}(x|\bm{\theta}^{(t+1)}, \bm{\zeta}^{(t+1)}) $ 
        
    }{
        \textcolor{gray}{\texttt{\# Optional further finetune the pruned subnet $\mathcal{F}(x|\hat{\bm{\theta}}, \bm{\zeta}^{(T_s)})$ with the original optimizer}}
    
        $ \bm{\theta}^{(t+1)} \leftarrow \bm{\theta}^{(t)} - \beta \frac{1}{n} \sum_{i=1}^{n} \nabla_{\bm{\theta}} \mathcal{L_{O}}(\bm{\theta}^{(t)})$ 
    }
    }
    
    \Return $ \mathcal{F^{\star}} \leftarrow \mathcal{F}_p(x|\bm{\theta}^{(T_r)})$

\end{algorithm}
\end{minipage}
\end{figure*}

\subsection{Unified and Progressive Pruning}
\label{Unified and Progressive Pruning}

\subsubsection{Unified Pruning}
\label{Unified Pruning}
The core idea of \textit{Unified Pruning} is to unifiedly instead of separately search on different modalities and structures, which enables adaptively instead of manually assigning the appropriate pruning ratio to each compressible component. Detailed implementation refers to Algorithm \ref{algorithm unified} in Appendix.

\vspace{-0.1cm}

\textbf{Unified Search on Different Modalities} \textit{Unified Pruning} groups the pruning masks by computation mechanisms. For typical vision-language Transformers, we divide the masks $\bm{\zeta} = \{\bm{\zeta}_{att}^{v},\ \bm{\zeta}_{att}^{l},\ \bm{\zeta}_{att}^{c}, \bm{\zeta}_{mlp}^{v},\ \bm{\zeta}_{mlp}^{l}\}$ into two groups:
\begin{equation}
    \setlength{\abovedisplayskip}{1pt}
    \setlength{\belowdisplayskip}{-3pt}
    \bm{\zeta}_a = \{\bm{\zeta}_{att}^{v},\ \bm{\zeta}_{att}^{l},\ \bm{\zeta}_{att}^{c} \},\quad \bm{\zeta}_m = \{\bm{\zeta}_{mlp}^{v},\ \bm{\zeta}_{mlp}^{l} \}.
\end{equation}

One group $\bm{\zeta}_a$ for different attention modules and another $\bm{\zeta}_m$ for different MLP modules. The ranking and selection of masks are performed within each group. Instead of searching on each $\bm{\zeta}_i \in \bm{\zeta}$ separately:
\begin{equation}
\setlength{\abovedisplayskip}{1pt}
\setlength{\belowdisplayskip}{-3pt}
\small
\quad \quad \bm{M}_i \leftarrow {\tt\small TopKMask}(\bm{\zeta}_i^{(T_s)}, \ p \cdot \text{Size}(\bm{\zeta}_i)) \ \ \text{for} \ \ \bm{\zeta}_i \in \bm{\zeta}, 
\end{equation}

where $\bm{M}_i$ is a binary mask used for pruning components of the subnet from the searched model. $\bm{M}_i$ is obtained by ranking and binarizing trainable mask $\bm{\zeta}_i$ at the final iteration $T_s$, which keeps the most important $p\%$ parameters. \textit{Unified Pruning} searches on different modalities within each group which ranks weights across different components:
\begin{equation}
\setlength{\abovedisplayskip}{0pt}
\setlength{\belowdisplayskip}{0pt}
\small
\bm{M}_{a} \leftarrow {\tt\small TopKMask}(\{{\bm{\zeta}_i^{(T_s)}} | \bm{\zeta}_i \in \ \bm{\zeta}_a\}, \ p \cdot \text{\small Size}(\bm{\zeta}_a)),
\end{equation}
\begin{equation}
\setlength{\belowdisplayskip}{-4pt}
\small
\bm{M}_{m} \leftarrow {\tt\small TopKMask}(\{{\bm{\zeta}_i^{(T_s)}} | \bm{\zeta}_i \in \ \bm{\zeta}_m\}, \ p \cdot \text{\small Size}(\bm{\zeta}_m)),
\end{equation}

% \vspace{-0.5cm}

\textbf{Unified Search on Different Structures}
We notice that simply uniting different structures degrades performance, and the reason is that the magnitude of the learned masks $\bm{\zeta}_i$ used for different structures vary greatly. 

Intuitively, it is feasible to conduct unified searching after transforming the magnitudes distributions of different structures' masks to have the same mean and standard deviation, and thus masks $\bm{\zeta}_i$ used for different structures can be comparable. For the simplicity of implementation, we individually transform the mean and standard deviation of magnitudes distributions of different structures' mask to the $0$ and $1$ by z-score standardization, respectively:
\begin{equation}
\setlength{\abovedisplayskip}{4pt}
\setlength{\belowdisplayskip}{4pt}
\small
\bm{\zeta}^{(T_s)}_a \leftarrow (\bm{\zeta}^{(T_s)}_a - \mathbb{E}[\bm{\zeta}^{(T_s)}_a]) / (\mathbb{E}[[\bm{\zeta}^{(T_s)}_a - \mathbb{E}[\bm{\zeta}^{(T_s)}_a]]^2])^{\frac{1}{2}},
\end{equation}
\begin{equation}
\setlength{\abovedisplayskip}{4pt}
\setlength{\belowdisplayskip}{-2pt}
\small
\bm{\zeta}^{(T_s)}_m \leftarrow (\bm{\zeta}^{(T_s)}_m - \mathbb{E}[\bm{\zeta}^{(T_s)}_m]) / (\mathbb{E}[[\bm{\zeta}^{(T_s)}_m - \mathbb{E}[\bm{\zeta}^{(T_s)}_m]]^2])^{\frac{1}{2}}.
\end{equation}

Then search on different modalities of different structures can be feasible:
\begin{equation}
\setlength{\abovedisplayskip}{2pt}
\setlength{\belowdisplayskip}{-4pt}
\small
    \bm{M} \leftarrow {\tt\small TopKMask}(\{{\bm{\zeta}_i^{(T_s)}} | \bm{\zeta}_i \in \ \bm{\zeta}\}, \ p \cdot \small{\tt\small Size}(\bm{\zeta})), 
\end{equation}

where $M$ is a binary mask used for pruning all compressible components, and $M$ is obtained by ranking and binarizing the whole trainable masks $\bm{\zeta}$ at the final iteration $T_s$.

\subsubsection{Progressive Pruning}
\label{Progressive Pruning}

Retrain the pruned model after the search is a traditional two-stage paradigm. However, this paradigm fails when it comes to high compression ratios, because there is no guarantee that the magnitude of searched mask $\bm{\zeta}^{(T_s)}$ corresponding to the eliminated neurons in compressible components will converge to $0$, which makes the pruned subnet with the parameters $\hat{\bm{\theta}}$ sliced from $\bm{\theta}^{(T_s)}$ difficult to converge. When the compression ratio becomes higher, the eliminated non-zero neurons from the parameters $\bm{\theta}^{(T_s)}$ is more, and the gap between $\hat{\bm{\theta}}$ and $\bm{\theta}^{(T_s)}$ is larger, thereby increasing the difficulty for the pruned subnet $\mathcal{F}(x|\hat{\bm{\theta}}, \bm{\zeta}^{(T_s)})$ to converge.

To address the above issue, we further propose the \textit{Progressive Pruning}, whose core idea is to ensure each magnitude of the trainable mask $\bm{\zeta}$ corresponding to the eliminated neurons in compressible components converges to $0$. This is achieved by updating trainable mask $\bm{\zeta}$ with a customed optimizer that is a function of the current iteration number $t$, instead of updating trainable mask $\bm{\zeta}$ with the same optimizer as the parameter $\bm{\theta}$ of the original model used.

Specifically, gradients $\bm{G}^{(t)}$ of $\bm{\zeta}$ in each iteration of the search phase is first collected:
\begin{equation}
\setlength{\abovedisplayskip}{0pt}
\setlength{\belowdisplayskip}{0pt}
\small
    \bm{G}^{(t)} \leftarrow \frac{1}{n} \sum_{i=1}^{n} \nabla_{\bm{\zeta}} \mathcal{L}(\bm{\theta}^{(t)}, \bm{\zeta}^{(t)}),
\end{equation}
where $n$ is the number of batch size. Then the accumulated gradients $\sum_{i=0}^{t}\ \bm{G}^{(i)}$ can be used as a new metric to evaluate the importance of corresponding neurons. And the pruning mask $\bm{M}^t$ at this iteration can be generated accordingly:
\begin{equation}
\setlength{\abovedisplayskip}{0pt}
\setlength{\belowdisplayskip}{0pt}
\small
    \bm{M}^t \leftarrow \small {\tt\small TopKMask}(\sum_{i=0}^{t} \bm{G}^{(i)}, \ p_t \cdot \small {\tt\small Size}(\bm{\zeta})),
\end{equation}
where $p_t$ is the current compression ratio when the iteration number is $t$. And the update strategy for optimizing $\bm{\zeta}$ in each iteration of the search phase can be written as
\begin{equation}
\setlength{\abovedisplayskip}{1pt}
\setlength{\belowdisplayskip}{1pt}
\small
    \bm{\zeta}^{(t+1)} \leftarrow (1 - \bm{M}_i^t) + (1 - \frac{p_t}{p})\bm{M}_i^t,
\end{equation}
which ensures that as $p_t$ progressively increases to $p$, each magnitude of mask $\bm{\zeta}$ corresponding to the removed neurons in compressible components will exactly converge to $0$. \textit{Progressive Pruning} eliminates the parameter gap between the searched model and the pruned subnet to be retrained, therefore gaining better convergence and performance, especially at high compression ratios.

The proposed \textit{UPop} framework combines \textit{Unified Pruning} and \textit{Progressive Pruning} as outlined in Algorithm \ref{algorithm Upop}. Line 2 $\sim$ 12 implements the search phase where Line 10 calculates the current compression ratio $p_t$ to be achieved (detailed discussion is provided in Appendix \ref{update p_t appendix}), and Line 13 $\sim$ 15 implements an optional retrain phase.

\begin{table*}[t]
  \setlength{\tabcolsep}{5pt}
  \captionsetup{font={small}}
  \setlength{\abovecaptionskip}{0.2cm}
  \small
  \begin{minipage}{0.6\linewidth}  
  \flushleft
  \caption{Compression results on the NLVR2. Bold indicates the best performance at the same compression ratio. Reduce indicates compression times. The marker \ \textcolor{ForestGreen}{\cmark} or \textcolor{brickred}{\xmark} \ indicates whether the model converges at the current compression times. The units of Params and FLOPs are M and G, respectively.}
  \begin{tabular}{c @{\hspace{1.0\tabcolsep}} c @{\hspace{1.0\tabcolsep}} c |c @{\hspace{1.0\tabcolsep}} c|l @{\hspace{1.0\tabcolsep}} l}
    \toprule
    Approach & Reduce & Status & Dev Acc & Test Acc & Params & FLOPs \\
    \midrule
    Uncompressed & $1 \times $ & \textcolor{ForestGreen}{\cmark} & 82.48 & 83.08 & 259.45 & 132.54 \\
    \midrule
    \multirow{2}{*}{\makecell{Mask-based \\ Pruning }} & $2 \times $ & \textcolor{ForestGreen}{\cmark} & 75.74 & 76.44 & 146.18 & 66.88 \\
    & $3 \times $ & \textcolor{brickred}{\xmark} & \textcolor{brickred}{\xmark} & \textcolor{brickred}{\xmark} & \textcolor{brickred}{\xmark} & \textcolor{brickred}{\xmark} \\
    \midrule
    & $2 \times $ & \textcolor{ForestGreen}{\cmark} & 79.50 & 80.32 & 149.90 & 95.01  \\
    & $3 \times $ & \textcolor{ForestGreen}{\cmark} & 71.25 & 71.66 & 106.33 & 68.19 \\
    \multirow{-3}{*}{\makecell{Unified \\ Pruning \\ (Ours) }} & $4 \times $ & \textcolor{brickred}{\xmark} & \textcolor{brickred}{\xmark} & \textcolor{brickred}{\xmark} & \textcolor{brickred}{\xmark} & \textcolor{brickred}{\xmark} \\
    \midrule
    & $2 \times $ & \textcolor{ForestGreen}{\cmark} & \textbf{80.33} & \textbf{81.13} & 150.15$_{\color{red}\downarrow 42\%}$ & 89.36$_{\color{red}\downarrow 33\%}$ \\
    & $3 \times $ & \textcolor{ForestGreen}{\cmark} & \textbf{76.89} & \textbf{77.61} & 109.01$_{\color{red}\downarrow 58\%}$ & 65.29$_{\color{red}\downarrow 51\%}$ \\
    & $4 \times $ & \textcolor{ForestGreen}{\cmark} & \textbf{72.85} & \textbf{73.55} & 88.61$_{\color{red}\downarrow 66\%}$ & 50.35$_{\color{red}\downarrow 62\%}$ \\
    & $5 \times $ & \textcolor{ForestGreen}{\cmark} & \textbf{68.71} & \textbf{68.76} & 76.81$_{\color{red}\downarrow 70\%}$ & 39.93$_{\color{red}\downarrow 70\%}$ \\
    \multirow{-5}{*}{\makecell{Unified and \\ Progressive \\ Pruning \\ (Ours) }} & $10 \times $ & \textcolor{ForestGreen}{\cmark} & \textbf{57.17} & \textbf{57.79} & 54.48$_{\color{red}\downarrow 79\%}$ & 19.08$_{\color{red}\downarrow 86\%}$ \\
    \bottomrule
  \label{table nlvr}
  \end{tabular}
  \vspace{-14pt}

  \end{minipage}
  \hspace{0.04\linewidth}
  % \medskip
  \begin{minipage}{0.36\linewidth}  
  \flushright
  \caption{Performance of the $2\times$ compressed BLIP model on the NLVR2 while searching only and without any retraining. The marker \textcolor{brickred}{\xmark} \ indicates the model fails converging.}
  \begin{tabular}{l | c @{\hspace{0.5\tabcolsep}} c @{\hspace{1.0\tabcolsep}}}
    \toprule
    Approach w/o Retrain & Dev Acc & Test Acc \\
    \midrule
    Mask-based Pruning & \textcolor{brickred}{\xmark} & \textcolor{brickred}{\xmark}  \\
    Unified Pruning (Ours) & \textcolor{brickred}{\xmark} & \textcolor{brickred}{\xmark} \\
    UPop (Ours) & \textbf{76.89} & \textbf{77.84} \\
    \bottomrule
  \end{tabular}
  \label{table search only}

  \vspace{7pt}
  \caption{Performance of the $2\times$ compressed BLIP model on the NLVR2 while retraining only one epoch after searching. }
  \begin{tabular}{l | c @{\hspace{0.5\tabcolsep}} c }
    \toprule
    Approach w/ Retrain & Dev Acc & Test Acc \\
    \midrule
    Mask-based Pruning & 62.82 & 63.35  \\
    Unified Pruning (Ours) & 75.42 & 75.30\\
    UPop (Ours) & \textbf{79.08} & \textbf{80.08} \\
    \bottomrule
  \end{tabular}
  \label{table one epoch retrain}

  \vspace{-3pt}
  \end{minipage}
  
\end{table*}

\begin{table*}[t]
  \setlength{\tabcolsep}{9pt}
  \captionsetup{font={small}}
  \small
  \centering
  \caption{Comparisons of the 2 $\times$ compressed BLIP model on the NLVR2. The superscript \textsuperscript{$*$}: The original approach is unstructured, which has the finest compression granularity, and therefore we report the performance at the same granularity as ours to achieve a fair comparison. The STE \textsuperscript{$\dagger$}: use straight-through estimator \cite{bengio2013estimating} to approximate gradients. }
  \begin{tabular}{l |c @{\hspace{1.0\tabcolsep}} c|l }
    \toprule
    Approach  & Dev Acc & Test Acc & Params \\
    \midrule
    Uncompressed \cite{li2022blip} & 82.48 & 83.08 & 259.45 \\
    \midrule
    Iterative Magnitude-based Pruning\textsuperscript{$*$} \cite{gan2022playing}  & 66.88 & 67.19 & 151.86 \\
    Mask-based Pruning \cite{chavan2022vision} & 75.74 & 76.44 & 146.18 \\
    Mask-based Pruning \cite{chavan2022vision} w/ Iterative Pruning with STE\textsuperscript{$\dagger$} \cite{sanh2020movement} & 78.05 & 77.68 & 146.11 \\ 
    \midrule
    Unified Pruning (Ours) & 79.50 & 80.32 & 149.90 \\
    Unified Pruning (Ours) w/ Iterative Pruning with STE\textsuperscript{$\dagger$} \cite{sanh2020movement} & 80.01 & 80.54 & 145.86 \\
    Unified Pruning (Ours) w/ Progressive Pruning (Ours) & \textbf{80.33} & \textbf{81.13} & 150.15 \\
    \bottomrule
  \end{tabular}
  \label{table nlvr cmp}
\end{table*}

\section{Experiments}
\label{EXPERIMENTS}

We report the performance of UPop on a series of multimodal tasks, including Visual Reasoning, Image Captioning, Visual Question Answer, and Image-Text Retrieval. Due to the space constraint, we provide more ablation studies and experiments on unimodal tasks in Appendix \ref{more exp}.

\subsection{Experiments on the Visual Reasoning Task}
\label{exp nlvr}
NLVR2 is a binary classification visual reasoning task with two images and a text description as inputs. To quantitatively evaluate the proposed UPop, we compress the fine-tuned BLIP model on this task at a ratio of 2, 3, 4, 5, and 10 times, respectively.\footnote{Note that at $N$ times compression, the total number of parameters will not be strictly equal to the $\frac{1}{N}$ of the original model. This is because some modules of the original model are not covered by the mask $\bm{\zeta}$, such as the patch embedding module, the word embedding module, and the classification head. In addition, at the same compression ratio, different searched masks will also lead to different structures and FLOPs of the compressed model.} The model consists of two weight-shared ViT as image encoder and a Bert with two cross-attention as text encoder, therefore the mask $\bm{\zeta}$ corresponding to the compressible components on this model is $\bm{\zeta} = \{\bm{\zeta}_{a}^{v},\ \bm{\zeta}_{m}^{v},\ \bm{\zeta}_{a}^{l},\ \bm{\zeta}_{m}^{l},\ \bm{\zeta}_{a}^{c0},\ \bm{\zeta}_{a}^{c1} \}$. As shown in Table \ref{table nlvr} and \ref{table nlvr cmp}, we compress the original model with Mask-based Pruning, Magnitude-based Pruning, Iterative Pruning, Unified Pruning, their combinations, and UPop, respectively.

\begin{table*}[t]
  \setlength{\tabcolsep}{4.25pt}
  \captionsetup{font={small}}
  \small
  \setlength{\abovecaptionskip}{0.2cm}
  \centering
  \caption{Comparisons between manually and adaptively assigning different compression ratios to different components of different modalities. Experiments are conducted with the 2 $\times$ compressed BLIP model on the NLVR2. The percentages in the table indicate the remaining parameters. Note that although the total compression ratio is the same for all experiments, the summation of individual percentages per row is different because the proportion of each component to the total number of model parameters is different. }
  \begin{tabular}{l |c @{\hspace{1.0\tabcolsep}} c @{\hspace{1.0\tabcolsep}} c @{\hspace{1.0\tabcolsep}} c @{\hspace{1.0\tabcolsep}} c @{\hspace{1.0\tabcolsep}} c |c }
    \toprule
    Assignment  & Vision Attention & Language Attention & Cross Attention1 & Cross Attention2 & Vision FFN & Language FFN & Test Acc \\
    \midrule
    Maunal& 50\% & 50\% & 50\% & 50\% & 50\% & 50\% & 76.44\\
    Maunal& 22\% & 62\% & 26\% & 75\% & 59\% & 53\% & 73.75\\
    Maunal& 44\% & 80\% & 17\% & 27\% & 83\% & 41\% & 80.01\\
    Maunal& 26\% & 73\% & 70\% & 63\% & 12\% & 64\% & 61.31\\
    Maunal& 58\% & 81\% & 19\% & 18\% & 47\% & 72\% & 78.80\\
    \midrule
    Adaptive (Ours)& 81\% & 58\% & 18\% & 19\% & 72\% & 47\% & \textbf{81.13} \\
    \bottomrule
  \end{tabular}
  \label{table assignment}
\end{table*}

\begin{figure*}[t]
    \centering
    \captionsetup{font={small}}
    \setlength{\abovecaptionskip}{0cm}
    \includegraphics[width=\textwidth]{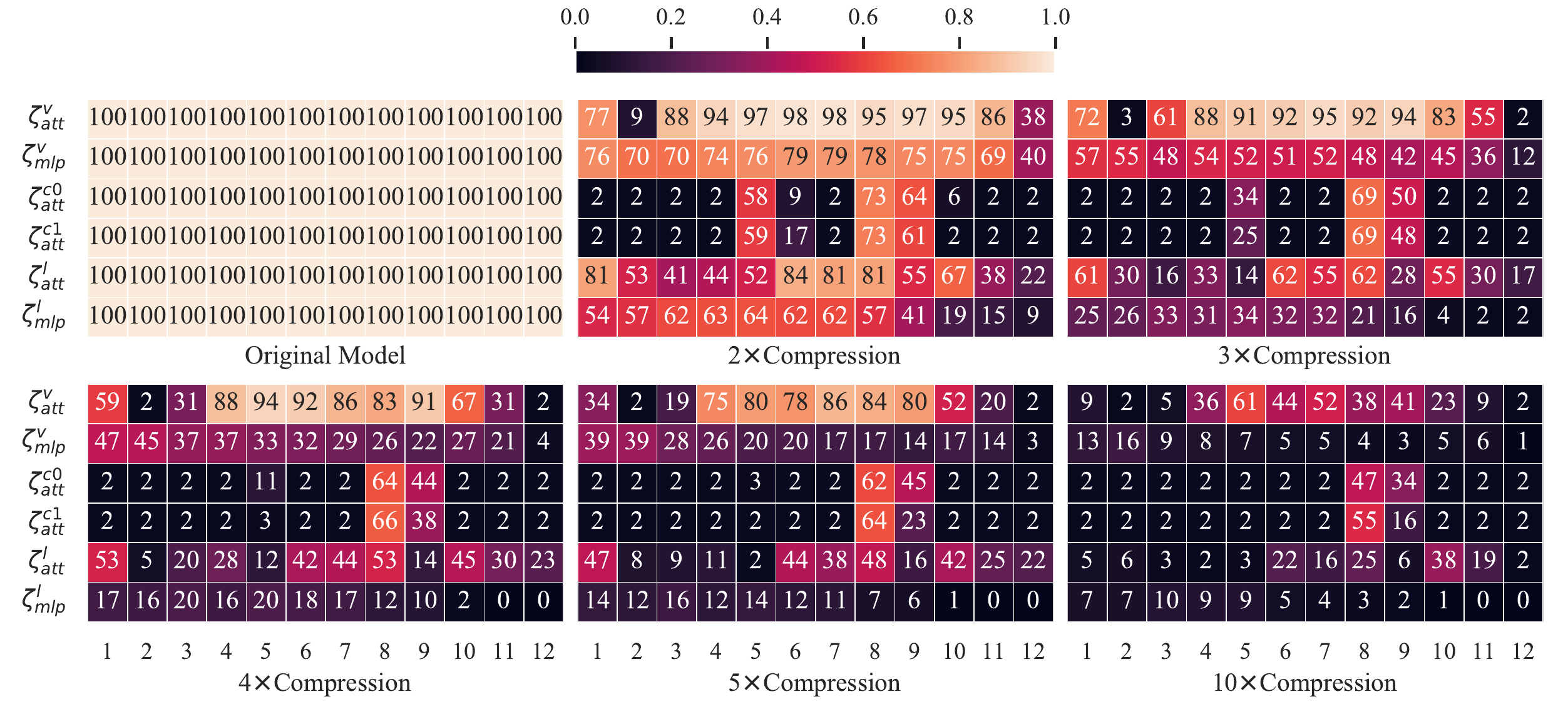}
    \vspace{-15pt}
    \caption{The proportion of all compressible components retained in the compressed BLIP model on the NLVR2. These six subfigures represent the original model and the compressed model at the $2 \times$, $3 \times$, $4 \times$, $5 \times$, and $10 \times$ compression ratio, respectively. In each subfigure, the horizontal axis represents the layer number, the vertical axis represents the compressible components corresponding to each $\bm{\zeta}_i$, and the number in cells represents the retained proportion of a certain component's certain layer.}
    \label{fig heatmap}
    \vspace{-12pt}
\end{figure*}

\subsection{Effect of Unified Pruning} 
\label{Effect of Unified Pruning}

At the $2 \times $ compression ratio, Table \ref{table nlvr} shows that compared to the Mask-based Pruning, Unified Pruning gains $3.76\%$ and $3.88\%$ accuracy improvement on the dev set and test set, respectively. Furthermore, Unified Pruning converges successfully at the $3 \times $ compression ratio, while Mask-based Pruning does not. Table \ref{table nlvr cmp} shows that Unified Pruning also outperforms Magnitude-based Pruning under the same setting of the compression ratio and granularity. 

Unified Pruning enables the model to adaptively assign appropriate compression ratios among different compressible components. Table \ref{table assignment} demonstrates that Unified Pruning can rescue us from the burden of repeated experiments (\eg, doing grid search) for searching the optimal compression ratio assignment. Furthermore, Figure \ref{fig heatmap} visualizes the proportion of all compressible components retained in the compressed model. It can be observed from the figure that retained proportion of each compressible component has significantly different trends as the compression ratio increases. Moreover, there are obviously unbalanced compression assignments in different layers at different compression ratios.

\begin{table*}[t]
  \setlength{\tabcolsep}{12 pt}
  \captionsetup{font={small}}
  \setlength{\abovecaptionskip}{0.2cm}
  \small
  \centering
  \caption{Compression results on the Image Caption task and the Visual Question Answering task. The CIDEr, SPICE, test-dev, and test-std are the higher the better. The units of Params and FLOPs are M and G, respectively.}
  \begin{tabular}{c @{\hspace{1.0\tabcolsep}} c|c @{\hspace{1.0\tabcolsep}} c|l @{\hspace{0.2\tabcolsep}} l @{\hspace{0.2\tabcolsep}} |c @{\hspace{1.0\tabcolsep}} c|l @{\hspace{0.2\tabcolsep}} l }
    \toprule
    \multirow{2}{*}{Approach} & \multirow{2}{*}{Reduce} & \multicolumn{4}{c|}{Image Caption}& \multicolumn{4}{c}{Visual Question Answering} \\
    \cmidrule{3-10}
    & & CIDEr & SPICE & Params & FLOPs & test-dev & test-std & Params & FLOPs \\
    \midrule
    Uncompressed & $1 \times $ & 133.3 & 23.8 & 224.0 & 65.7 & 77.4 & 77.5 & 361.6 & 186.1 \\
    \midrule
    \multirow{2}{*}{\makecell{Mask-based \\ Pruning}} & $2 \times $ & 112.9 & 21.0 & 124.9 & 33.2 & 71.6 & 71.6 & 205.8 & 96.4 \\
    & $4 \times $ & 60.7 & 12.8 & 75.4 & 17.1 & 69.2 & 69.3 & 128.4 & 51.7  \\
    \midrule
    & $2 \times $  & 127.9 & 23.1 & 124.7 & 44.2 & 75.2 & 75.4 & 216.4 & 118.7 \\
    \multirow{-2}{*}{\makecell{Unified Pruning \\ (Ours)}} & $4 \times $ & 100.3 & 19.1 & 77.5 & 25.6 & 73.5 & 73.6 & 135.3 & 77.3 \\
    \midrule
    & $2 \times $ & \textbf{128.9} & \textbf{23.3} & 127.1$_{\color{red}\downarrow 43\%}$ & 39.8$_{\color{red}\downarrow 39\%}$ & \textbf{76.3} & \textbf{76.3} & 211.3$_{\color{red}\downarrow 42\%}$ & 109.4$_{\color{red}\downarrow 41\%}$ \\
    \multirow{-2}{*}{\makecell{UPop \\ (Ours) }} & $4 \times $ & \textbf{117.4} & \textbf{21.7} & 76.5$_{\color{red}\downarrow 66\%}$ & 22.2$_{\color{red}\downarrow 66\%}$ & \textbf{74.5} & \textbf{74.6} & 133.3$_{\color{red}\downarrow 63\%}$ & 62.3$_{\color{red}\downarrow 67\%}$ \\
    \bottomrule
    \label{table Image Caption and VQA}
  \end{tabular}
  \vspace{-14pt}
\end{table*}

\begin{table*}[!htbp]
  \setlength{\tabcolsep}{11.5 pt}
  \captionsetup{font={small}}
  \setlength{\abovecaptionskip}{0.2cm}
  \small
  \centering
  \caption{Compress BLIP on the COCO and Flickr30K datasets of the Image-Text Retrieval task. The R@1, R@5, and R@10 are the higher the better. The units of Params and FLOPs are M and G, respectively.}
  \begin{tabular}{c @{\hspace{1.0\tabcolsep}} c @{\hspace{1.0\tabcolsep}} c|c @{\hspace{1.0\tabcolsep}} c @{\hspace{1.0\tabcolsep}} c|c @{\hspace{1.0\tabcolsep}} c @{\hspace{1.0\tabcolsep}} c|l @{\hspace{0.2\tabcolsep}} l}
    \toprule
    \multirow{2}{*}{Dataset} & \multirow{2}{*}{Approach} & \multirow{2}{*}{Reduce} & \multicolumn{3}{c}{Image $\rightarrow$ Text} & \multicolumn{3}{c|}{Text $\rightarrow$ Image} & \multirow{2}{*}{Params} & \multirow{2}{*}{FLOPs} \\ 
    \cmidrule{4-9}
    & & & R@1 & R@5 & R@10 & R@1 & R@5 & R@10 &  \\
    \midrule
    \multirow{7}{*}{\makecell{COCO\\(5K test set)}} & Uncompressed & $1 \times $ & 81.9 & 95.4 & 97.8 & 64.3 & 85.7 & 91.5 & 447.6 & 153.2 \\
    \cmidrule{2-11}
    & \multirow{2}{*}{\makecell{Mask-based \\ Pruning}} & $2 \times $ & 61.7 & 85.0 & 91.1 & 46.0 & 73.2 & 82.6 & 249.5 & 77.3\\
    & & $4 \times $ & \textcolor{brickred}{\xmark} & \textcolor{brickred}{\xmark} & \textcolor{brickred}{\xmark} & \textcolor{brickred}{\xmark} & \textcolor{brickred}{\xmark} & \textcolor{brickred}{\xmark} & \textcolor{brickred}{\xmark} & \textcolor{brickred}{\xmark} \\
    \cmidrule{2-11}
    & \multirow{2}{*}{\makecell{Unified Pruning \\ (Ours)}} & $2 \times $ & 75.4 & 92.9 & 96.3 & 57.6 & 81.9 & 88.7 & 253.1 & 103.4 \\
    & & $4 \times $ & 40.3 & 69.3 & 80.2 & 31.3 & 58.8 & 70.7 & 148.7 & 61.4\\
    \cmidrule{2-11}
    & \multirow{2}{*}{\makecell{UPop \\ (Ours)}} & $2 \times $ & \textbf{77.4} & \textbf{93.4} & \textbf{97.0} & \textbf{59.8} & \textbf{83.1} & \textbf{89.8} & 248.9$_{\color{red}\downarrow 44\%}$ & 88.3$_{\color{red}\downarrow 42\%}$ \\
    & & $4 \times $ & \textbf{62.9} & \textbf{86.2} & \textbf{92.3} & \textbf{47.4} & \textbf{74.8} & \textbf{83.9} & 147.9$_{\color{red}\downarrow 67\%}$ & 50.2$_{\color{red}\downarrow 67\%}$\\
    \cmidrule{1-11}
    \multirow{7}{*}{\makecell{Flickr30K\\(1K test set)}} & Uncompressed & $1 \times $ & 96.8 & 99.9 & 100.0 & 86.9 & 97.3 & 98.7 & 447.6 & 153.2 \\
    \cmidrule{2-11}
    & \multirow{2}{*}{\makecell{Mask-based \\ Pruning}} & $2 \times $ & 78.9 & 92.7 & 95.5 & 63.8 & 85.1 & 90.1 & 249.3 & 77.2\\
    & & $4 \times $ & \textcolor{brickred}{\xmark} & \textcolor{brickred}{\xmark} & \textcolor{brickred}{\xmark} & \textcolor{brickred}{\xmark} & \textcolor{brickred}{\xmark} & \textcolor{brickred}{\xmark} & \textcolor{brickred}{\xmark} & \textcolor{brickred}{\xmark} \\
    \cmidrule{2-11}
    & \multirow{2}{*}{\makecell{Unified Pruning \\ (Ours)}} & $2 \times $ & 92.2 & 99.0 & \textbf{99.8} & 78.5 & 93.7 & 96.1 & 252.3 & 104.1 \\
    & & $4 \times $ & 50.0 & 76.1 & 84.3 & 40.8 & 68.1 & 77.0 & 148.7 & 60.8 \\
    \cmidrule{2-11}
    & \multirow{2}{*}{\makecell{UPop \\ (Ours)}} & $2 \times $ & \textbf{94.0} & \textbf{99.5} & 99.7 & \textbf{82.0} & \textbf{95.8} & \textbf{97.6} & 250.5$_{\color{red}\downarrow 44\%}$ & 91.0$_{\color{red}\downarrow 41\%}$ \\
    & & $4 \times $ & \textbf{85.8} & \textbf{97.4} & \textbf{98.4} & \textbf{71.3} & \textbf{91.0} & \textbf{94.8} & 147.6$_{\color{red}\downarrow 67\%}$ & 51.0$_{\color{red}\downarrow 67\%}$ \\
    \bottomrule
  \label{table BLIP Image-Text Retrieval}
  \vspace{-10pt}
  \end{tabular}
\end{table*}

\subsection{Effect of Progressive Pruning} 
\label{Effect of Progressive Pruning}

As shown in Table \ref{table nlvr}, at the $2 \times $ compression ratio, the Unified and Progressive Pruning (UPop) gains further $0.83\%$ and $0.81\%$ accuracy improvement on the dev set and test set compared to the Unified Pruning. Moreover, at the $3 \times $ compression, the improvements are extended to $5.64\%$ and $5.95\%$, respectively. At the higher $4 \times $, $5 \times $, and $10 \times $ compression ratio, the Progressive Pruning can still enable the compressed model to converge successfully, while both Mask-based Pruning and Unified Pruning fail. Furthermore, Table \ref{table nlvr cmp} shows that the proposed  Progressive Pruning also outperforms Iterative Pruning with STE under the same setting of the compression ratio and granularity.

To further illustrate how Progressive Pruning strengthens the convergence capability of the compressed model, we compare the performance of pruned subnets in the situation that search without any retraining or search with only one epoch retraining. Table \ref{table search only} shows that the model compressed by UPop can converge without any retraining while the other two compression approaches fail. Furthermore, Table \ref{table one epoch retrain} shows that with only one epoch retraining, the model compressed by UPop converges at significantly superior performance to the other two approaches. The experiments in Table \ref{table search only} and \ref{table one epoch retrain} indicate that Progressive Pruning maintains the convergence capability of the compressed model by initializing the pruned subnet to be retrained with better parameter weights.

\begin{table*}[t]
  \setlength{\tabcolsep}{12 pt}
  \captionsetup{font={small}}
  \setlength{\abovecaptionskip}{0.2cm}
  \small
  \centering
  \caption{Compress CLIP on the COCO and Flickr30K datasets of the Image-Text Retrieval task. Notations are the same as in Table \ref{table BLIP Image-Text Retrieval}.}
  \begin{tabular}{c @{\hspace{1.0\tabcolsep}} c @{\hspace{1.0\tabcolsep}} c|c @{\hspace{1.0\tabcolsep}} c @{\hspace{1.0\tabcolsep}} c|c @{\hspace{1.0\tabcolsep}} c @{\hspace{1.0\tabcolsep}} c|l @{\hspace{0.2\tabcolsep}} l @{\hspace{0.2\tabcolsep}} }
    \toprule
    \multirow{2}{*}{Dataset} & \multirow{2}{*}{Approach} & \multirow{2}{*}{Reduce} & \multicolumn{3}{c}{Image $\rightarrow$ Text} & \multicolumn{3}{c|}{Text $\rightarrow$ Image} & \multirow{2}{*}{Params} & \multirow{2}{*}{FLOPs} \\ 
    \cmidrule{4-9}
    & & & R@1 & R@5 & R@10 & R@1 & R@5 & R@10 &  \\
    \midrule
    \multirow{3}{*}{\makecell{COCO\\(5K test set)}} & Uncompressed & $1 \times $ & 71.5 & 90.8 & 95.4 & 56.8 & 80.7 & 87.6 & 856.0 & 395.7 \\
    \cmidrule{2-11}
    & & $2 \times $ & 70.8 & 90.8 & 95.2 & 53.1 & 79.9 & 87.3 & 473.7$_{\color{red}\downarrow 45\%}$ & 196.3$_{\color{red}\downarrow 50\%}$ \\
    & \multirow{-2}{*}{\makecell{UPop \\ (Ours)}} & $4 \times $ & 56.1 & 82.4 & 90.2 & 41.1 & 71.0 & 81.4 & 280.2$_{\color{red}\downarrow 67\%}$ & 105.9$_{\color{red}\downarrow 73\%}$ \\
    \cmidrule{1-11}
    \multirow{3}{*}{\makecell{Flickr30K\\(1K test set)}} & Uncompressed & $1 \times $ & 96.8 & 100.0 & 100.0 & 86.6 & 97.8 & 99.1 & 856.0 & 395.7 \\
    \cmidrule{2-11}
    & & $2 \times $ & 93.2 & 99.4 & 99.8 & 80.5 & 95.4 & 97.6 & 474.3$_{\color{red}\downarrow 45\%}$ & 201.1$_{\color{red}\downarrow 49\%}$ \\
    & \multirow{-2}{*}{\makecell{UPop \\ (Ours)}} & $4 \times $ & 82.9 & 95.7 & 97.8 & 67.3 & 89.5 & 93.5 & 278.5$_{\color{red}\downarrow 67\%}$ & 102.6$_{\color{red}\downarrow 74\%}$ \\
    \bottomrule
  \label{table CLIP Image-Text Retrieval}
  \vspace{-10pt}
  \end{tabular}
\end{table*}

\begin{figure*}[t]
    \centering
    \captionsetup{font={small}}
    \setlength{\abovecaptionskip}{0cm}
    \includegraphics[width=\textwidth]{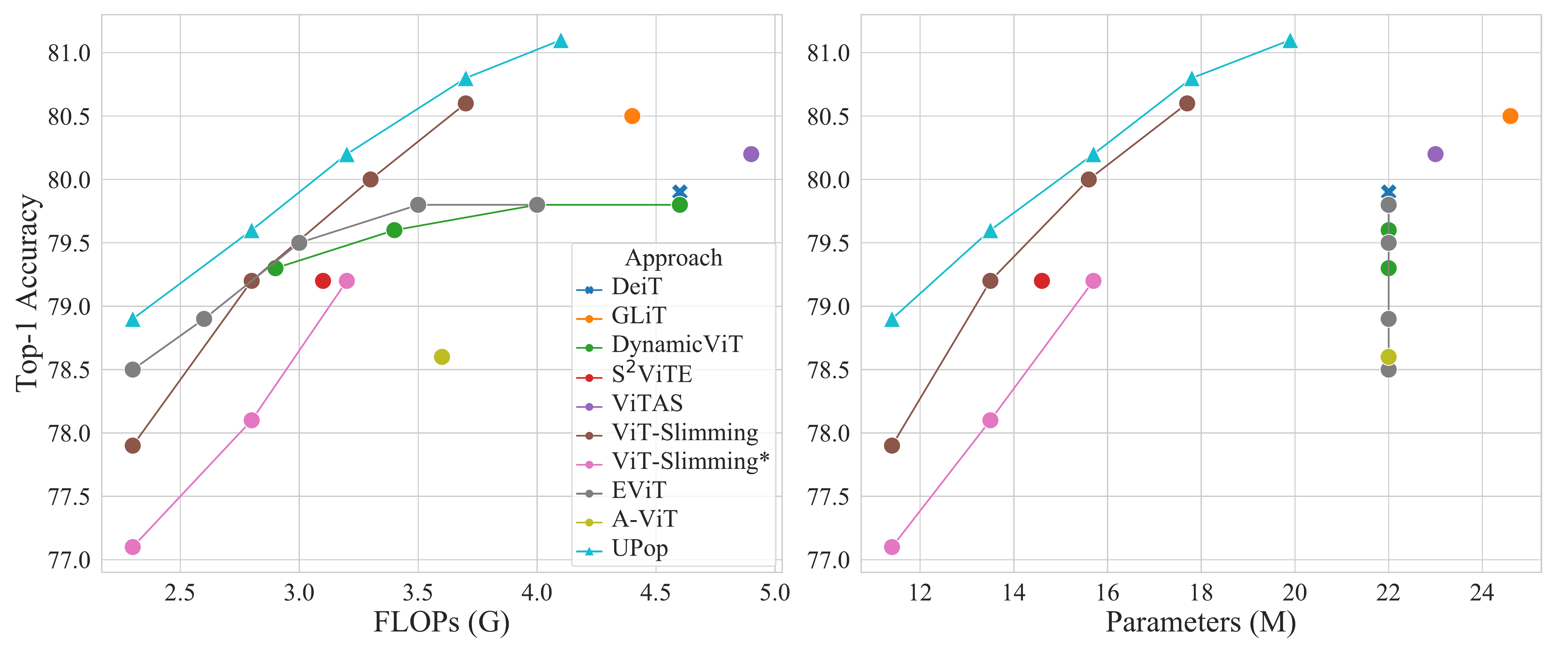}
    \caption{The left and right subfigures illustrate the Accuracy-FLOPs and Accuracy-Parameter trade-off, respectively. \textsuperscript{$*$} indicates the performance of the deployable model if the original model is non-deployable. Two subfigures demonstrate that the proposed UPop (marked with the blue triangle) achieves better performance on both trade-offs. Note that token-specific compression approaches only reduce FLOPs and not the number of parameters. Therefore they are vertical lines in the Accuracy-Parameter trade-off figure. }
    \label{fig FLOPsAndParams}
    \vspace{-10pt}
\end{figure*}

\subsection{Experiments on the Image Caption Task}
\label{exp caption}
To validate the versatility of the proposed UPop, we further conducted experiments on the Image Caption task. We compress the fine-tuned BLIP model on the COCO dataset at a ratio of 2 and 4 times, respectively. The model consists of a ViT as the image encoder and a Bert with cross-attention as the text decoder. Therefore the mask $\bm{\zeta}$ corresponding to the compressible components on this model is $\bm{\zeta} = \{\bm{\zeta}_{a}^{v},\ \bm{\zeta}_{m}^{v},\ \bm{\zeta}_{a}^{l},\ \bm{\zeta}_{m}^{l},\ \bm{\zeta}_{a}^{c} \}$. Table \ref{table Image Caption and VQA} shows that UPop also achieves superior performance on the Image Caption task.

\subsection{Experiments on the Visual QA Task}
\label{exp vqa}

We compress the fine-tuned BLIP model on the VQA2.0 dataset at a ratio of 2 and 4 times, respectively. The model consists of a ViT as the image encoder, a Bert with cross-attention as the text encoder, and a Bert with cross-attention as the text decoder. Therefore the mask $\bm{\zeta}$ corresponding to the compressible components on this model is $\bm{\zeta} = \{\bm{\zeta}_{a}^{v},\ \bm{\zeta}_{m}^{v},\ \bm{\zeta}_{a}^{l,en},\ \bm{\zeta}_{m}^{l,en},\ \bm{\zeta}_{a}^{l,de},\ \bm{\zeta}_{m}^{l,de}\}$. Table \ref{table Image Caption and VQA} shows the improved performance of UPop on the VQA task.

\subsection{Experiments on the Retrieval Task}
\label{exp retrieval}

We compress the fine-tuned BLIP model on the COCO and Flickr30K datasets at a ratio of 2 and 4 times, respectively. The model consists of a ViT as the image encoder, a Bert with cross-attention as the text encoder, an extra ViT as the momentum image encoder, and an extra Bert with cross-attention as the momentum text encoder. Since the momentum models are updated by taking the moving average of normal models, we do not add the compression mask into the momentum models. Therefore the mask $\bm{\zeta}$ corresponding to the compressible components on this model is $\bm{\zeta} = \{\bm{\zeta}_{a}^{v},\ \bm{\zeta}_{m}^{v},\ \bm{\zeta}_{a}^{l},\ \bm{\zeta}_{m}^{l},\ \bm{\zeta}_{a}^{c} \}$. Table \ref{table BLIP Image-Text Retrieval} shows the improved performance of UPop on the Image-Text Retrieval task.

To further validate the versatility of UPop on different model architectures, we also compressed the dual-stream architecture, CLIP \cite{radford2021learning}, on the Image-Text Retrieval task. Table \ref{table CLIP Image-Text Retrieval} shows that UPop is able to achieve comparable effectiveness to BLIP on CLIP. \footnote{Note that we use the momentum distillation to finetune CLIP on the Image-Text Retrieval task. Due to the introduction of momentum models, the number of parameters and FLOPs in Table \ref{table CLIP Image-Text Retrieval} are approximately twice as high as the original CLIP, respectively.}

\subsection{Experiments on the Image Classification Task}
\label{exp classification}
In addition to the multimodal tasks that UPop mainly focuses on, UPop can also be adapted to unimodal tasks by combining Unified Search on different structures and Progressive Pruning. As illustrated in Figure \ref{fig FLOPsAndParams} and reported in Appendix Table \ref{table imagenet}, we conduct DeiT \cite{touvron2021training} compression on ImageNet dataset \cite{deng2009imagenet}, and UPop can also achieve competitive performance compared to other unimodal compression SOTA approaches.

\section{Conclusion}
This paper proposes a novel multimodal pruning framework, Unified and Progressive Pruning (UPop), for vision-language Transformers. UPop unifiedly searches on all compressible components consisting of Self-Attentions, MLPs, and Cross-Attentions of all modalities, and thus can adaptively assign appropriate compression ratios for all components. Moreover, analysis of masks indicates that the importance of components for compression varies. Therefore, the proposed unified search is a better choice than manually assigning compression ratios among different components, which is inefficient and sub-optimal. Furthermore, UPop progressively conducts search and retraining, which effectively strengthens the convergence capability of the compressed model and enables higher compression ratios.

\section{Acknowledgements}

This work was supported by the National Key R$\&$D Program of China (2022YFB4701400/4701402), SZSTC Grant (JCYJ20190809172201639, WDZC20200820200655001), Shenzhen Key Laboratory (ZDSYS20210623092001004), Beijing Key Lab of Networked Multimedia, and Shanghai AI Laboratory.

\newpage
\bibliography{main}
\bibliographystyle{icml2023}

%%%%%%%%%%%%%%%%%%%%%%%%%%%%%%%%%%%%%%%%%%%%%%%%%%%%%%%%%%%%%%%%%%%%%%%%%%%%%%%
%%%%%%%%%%%%%%%%%%%%%%%%%%%%%%%%%%%%%%%%%%%%%%%%%%%%%%%%%%%%%%%%%%%%%%%%%%%%%%%
% APPENDIX
%%%%%%%%%%%%%%%%%%%%%%%%%%%%%%%%%%%%%%%%%%%%%%%%%%%%%%%%%%%%%%%%%%%%%%%%%%%%%%%
%%%%%%%%%%%%%%%%%%%%%%%%%%%%%%%%%%%%%%%%%%%%%%%%%%%%%%%%%%%%%%%%%%%%%%%%%%%%%%%
\newpage
\appendix
\onecolumn

\label{Appendix}

\section{Implementation Details}

\subsection{Hyperparameter Settings}

\medskip

\begin{table*}[!htbp]
  \captionsetup{font={small}}
  \setlength{\abovecaptionskip}{0.2cm}
  \small
  \centering
  \begin{minipage}{\linewidth}
  \setlength{\tabcolsep}{9 pt}
  \caption{Training hyperparameters for compressing BLIP-based models.}
  \begin{tabular}{l @{\hspace{1.0\tabcolsep}} c @{\hspace{1.0\tabcolsep}} c c @{\hspace{1.5\tabcolsep}} c @{\hspace{1.5\tabcolsep}} c}
    \toprule
    \multirow{4}{*}{Hyperparameters} & \multirow{2}{*}{\makecell{BLIP-NLVR \\ \cite{li2022blip}}} & \multirow{2}{*}{\makecell{BLIP-Caption\\ \cite{li2022blip}}} & \multirow{2}{*}{\makecell{BLIP-VQA\\ \cite{li2022blip}}} & \multicolumn{2}{c}{\multirow{2}{*}{\makecell{BLIP-Retrieval\\ \cite{li2022blip}}}}  \\ 
    & & & & & \\
    \cmidrule{2-6}
    & \multirow{2}{*}{\makecell{NLVR2\\ \cite{suhr2018corpus}}} & \multirow{2}{*}{\makecell{COCO\\ \cite{lin2014microsoft}}} & \multirow{2}{*}{\makecell{VQAv2\\ \cite{goyal2017making}}} & \multirow{2}{*}{\makecell{COCO\\ \cite{lin2014microsoft}}} & \multirow{2}{*}{\makecell{Flickr30K\\ \cite{young2014image}}} \\
    & & & & & \\
    \midrule
    Optimizer & \multicolumn{5}{c}{AdamW \cite{loshchilov2017decoupled}} \\
    AdamW $\beta$ & \multicolumn{5}{c}{(0.9, 0.999)} \\
    Weight decay & \multicolumn{5}{c}{0.05} \\
    Batch size & \multicolumn{5}{c}{256} \\
    Search epochs & 15 & 5 & 10 & 6 & 12 \\
    Search LR & 3e-6 & 1e-5 & 2e-5 & 1e-5 & 1e-5 \\
    Retrain epochs & 15 & 5 & 10 & 6 & 12 \\
    Retrain LR & 3e-6 & 1e-5 & 2e-5 & 1e-5 & 1e-5 \\
    Search LR schedule & \multicolumn{5}{c}{N/A} \\
    Retrain LR schedule & \multicolumn{5}{c}{CosineLRScheduler \cite{loshchilov2016sgdr}} \\
    Data augmentation & \multicolumn{5}{c}{RandomAugment \cite{Cubuk_2020_CVPR_Workshops}}\\
    \bottomrule
  \label{table training hyperparameters}
  \end{tabular}
  \end{minipage}

  \medskip
  \bigskip
  
  \begin{minipage}{\linewidth}
  \setlength{\tabcolsep}{7 pt}
  \caption{Training hyperparameters for compressing CLIP, DeiT, and Segmenter.}
  \begin{tabular}{l @{\hspace{1.0\tabcolsep}} c @{\hspace{1.0\tabcolsep}} c @{\hspace{1.0\tabcolsep}} c @{\hspace{1.0\tabcolsep}} c}
    \toprule
    \multirow{3}{*}{Hyperparameters} & \multicolumn{2}{c}{CLIP \cite{radford2021learning}} & DeiT \cite{touvron2021training} & Segmenter \cite{strudel2021segmenter} \\ 
    \cmidrule{2-5}
    & \multirow{2}{*}{\makecell{COCO\\ \cite{lin2014microsoft}}} & \multirow{2}{*}{\makecell{Flickr30K\\ \cite{young2014image}}} & \multirow{2}{*}{\makecell{ImageNet\\ \cite{deng2009imagenet}}} & \multirow{2}{*}{\makecell{ADE20k\\ \cite{zhou2017scene}}} \\
    & & & & \\
    \midrule
    Optimizer & \multicolumn{3}{c}{AdamW \cite{loshchilov2017decoupled}} & SGD \cite{robbins1951stochastic} \\
    Optimizer settings  & \multicolumn{3}{c}{AdamW $\beta$ (0.9, 0.999)} & SGD momentum 0.9\\
    Weight decay & 0.2 & 0.2 & 0.05 & 0 \\
    Batch size & 256 & 256 & 4096 & 64 \\
    Search epochs & 6 & 12 & 60 & 16 \\
    Search LR & 1e-5 & 1e-5 & 8e-4 & 4e-3 \\
    Retrain epochs & 6 & 12 & 300 & 64 \\
    Retrain LR & 1e-5 & 1e-5 & 8e-4 & 4e-3 \\
    Search LR schedule & \multicolumn{4}{c}{N/A} \\
    Retrain LR schedule & \multicolumn{3}{c}{CosineLRScheduler \cite{loshchilov2016sgdr}} & PolynomialLR \cite{strudel2021segmenter} \\
    Data augmentation & \multicolumn{2}{c}{\makecell{RandomAugment\\ \cite{Cubuk_2020_CVPR_Workshops}}} & \makecell{RepeatedAugment\\ \cite{touvron2021training}} & \makecell{Standard pipline from \\ MMSegmentation \cite{contributors2020mmsegmentation}} \\
    \bottomrule
  \label{table training hyperparameters}
  \end{tabular}
  \end{minipage}

  \medskip
  \bigskip
   
  \begin{minipage}{\linewidth}
  \setlength{\tabcolsep}{7 pt}
  \caption{Structure hyperparameters for all models used in our experiments. The superscript \textsuperscript{$*$} indicates 2 Transformers share parameters. The superscript \textsuperscript{$\dagger$}: 12 layers for the encoder and 2 layers for the decoder.}
  \begin{tabular}{lccccccccc}
    \toprule
    \multirow{2}{*}{Model} & \multirow{2}{*}{\makecell{Input\\resolution}} & \multicolumn{4}{c}{Vision Transformer} & \multicolumn{4}{c}{Language Transformer} \\
    & & number & layers & width & heads & number & layers & width & heads \\
    \midrule
    BLIP-NLVR \cite{li2022blip}& 384$\times$384 & 2\textsuperscript{$*$} & 12 & 768 & 12 & 1 & 12 & 768 & 12 \\
    BLIP-Caption \cite{li2022blip}& 384$\times$384 & 1 & 12 & 768 & 12 & 1 & 12 & 768 & 12 \\
    BLIP-VQA \cite{li2022blip}& 480$\times$480 & 1 & 12 & 768 & 12 & 2 & 12 & 768 & 12 \\
    BLIP-Retrieval \cite{li2022blip} & 384$\times$384 & 2 & 12 & 768 & 12 & 2 & 12 & 768 & 12 \\
    CLIP \cite{radford2021learning} & 336$\times$336 & 2 & 24 & 1024 & 16 & 2 & 12 & 768 & 12 \\
    DeiT \cite{touvron2021training} & 224$\times$224 & 1 & 12 & 384 & 6 & 0 & - & - & - \\
    Segmenter \cite{strudel2021segmenter} & 512$\times$512 & 2 & (12, 2)\textsuperscript{$\dagger$} & 384 & 6 & 0 & - & - & - \\
    \bottomrule
  \label{table structure hyperparameters}
  \end{tabular}
  \end{minipage}
  
\end{table*}

\newpage

\subsection{Scope of Compressible Components}

Self-Attentions, Cross-Attentions, and MLPs are widely used components in multimodal transformer layers. Consequently, the scope of compressible components in our experiments includes Self-Attentions, MLPs, and Cross-Attentions of both Vision Transformers and Language Transformers. Note that Cross-Attention only needs to be compressed if it exists. In early multimodal Transformers, \eg, LXMERT \cite{tan2019lxmert} and ViLBERT \cite{lu2019vilbert}, Cross-Attention exists within both vision and language Transformers. In some more modern works, Cross-Attention exists in only one of the modalities, such as CoCa \cite{yu2022coca} and BLIP \cite{li2022blip}. In addition, there are also a few models, such as CLIP \cite{radford2021learning}, that do not have explicit Cross-Attention but only conduct cross-modality interaction by maximizing the cosine similarity of outputs from different modalities.

\subsection{Compression Granularity and Deployability}
UPop is a structured pruning approach whose minimum granularity is an entire row or column in the weights of model parameters, and a deployable pruning approach that allows the compressed model to be physically extracted from the original model. Generally speaking, structured approaches such as UPop are relatively easier to deploy, while unstructured approaches such as \cite{gan2022playing} are relatively hard to deploy. More specifically, suppose we are going to prune a fully connected layer with the parameter $\bm{\theta} \in \mathbb{R}^{w_1 \times w_2}$, then unstructured approaches will add a binary mask of the same shape $w_1 \times w_2$ on the $\bm{\theta}$,  and every row and column of the mask may have a different amount of 0 after pruning, which results in difficulty for extracting all the weights with a mask of 1 to constitute a legal and smaller parameter matrix for the pruned layer. However, UPop will add a binary mask of the shape $w_1$ on the output of the fully connected layer. Hence each position in the mask of UPop corresponds to an entire row of the parameter matrix $\bm{\theta}$. It is simple to constitute a legal and smaller parameter matrix for the pruned layer by physically removing the entire rows from the original parameter matrix $\bm{\theta}$. 

Besides, ViT-Slimming \cite{chavan2022vision} compress heads of Self-Attentions with unrestricted compression ratio, and thus the compressed model may have different embedding sizes of heads within a layer. However, the matrix computation of the attention map on regular hardware (\eg, GPU cards) requires the query and key of each head within a layer to have the same embedding size. By restricting each head within the same layer to have the same compression ratio, UPop frees from non-deployable matrix computation and becomes structured across heads within individual layers.

\subsection{Implentation of Mask-based Pruning}
\label{IMPLEMENTATION OF Mask-based Pruning}

The \textit{Mask-based Pruning} is outlined in Algorithm \ref{algorithm mask-based}. Line 1 $\sim$ 10 implements the search phase, and Line 11 $\sim$ 13 implements the retrain phase.

\begin{figure}[ht]
  \centering
  \vspace{-10pt}
\begin{minipage}[b]{\linewidth}
\begin{algorithm}[H]
    % \small
    \caption{Mask-based Pruning}
    \label{algorithm mask-based}
    \setcounter{AlgoLine}{0}
    \LinesNumbered
    
    \KwIn{Original model $\mathcal{F}$, parameters of the original model $\bm{\theta}$, parameters of the trainable mask $\bm{\zeta}$, total compression ratio $p$, iterations in the search stage $T_{s}$ and retrain stage $T_{r}$, learning rate for the search stage $\alpha$ and retrain stage $\beta$ }

    \KwOut{Model $\mathcal{F^{\star}}$ after the search and retrain}
    
    \For{$t\gets0$ \KwTo $T_{s} - 1$} {

        \textcolor{gray}{\texttt{\# Calculate the loss $\mathcal{L}$}}
        
        $\mathcal{L} \leftarrow \mathcal{L_{O}} + w_{a} \sum_{\bm{\zeta}_i \in \bm{\zeta}_a} \lVert \bm{\zeta}_i \rVert_{1} + w_{m} \sum_{\bm{\zeta}_i \in \bm{\zeta}_m} \lVert \bm{\zeta}_i \rVert_{1}$ 

        \textcolor{gray}{\texttt{\# Normally update both $\bm{\theta}$ and $\bm{\zeta}$ with the original optimizer}}
        
        $ \bm{\theta}^{(t+1)} \leftarrow \bm{\theta}^{(t)} - \alpha \frac{1}{n} \sum_{i=1}^{n} \nabla_{\bm{\theta}} \mathcal{L}(\bm{\theta}^{(t)}, \bm{\zeta}^{(t)})$, \quad 
        $ \bm{\zeta}^{(t+1)} \leftarrow \bm{\zeta}^{(t)} - \alpha \frac{1}{n} \sum_{i=1}^{n} \nabla_{\bm{\zeta}} \mathcal{L}(\bm{\theta}^{(t)}, \bm{\zeta}^{(t)})$ 
    }

    \textcolor{gray}{\texttt{\# Individually generate each pruning mask $\bm{M}_i$ by ranking and selecting on each $\bm{\zeta}_i$}}

    \For{$\bm{\zeta}_i \in \bm{\zeta}$}{ 
    
        $\bm{M}_i \leftarrow \small {\tt TopKMask}(\bm{\zeta}_i^{(T_s)}, \ p \cdot \small {\tt Size}(\bm{\zeta}_i))$
    }

    \textcolor{gray}{\texttt{\# Compress each $\bm{\theta}_i$ based on each $M_i$ and accordingly compress $\mathcal{F}$}}
    
    $ \hat{\bm{\theta}} \leftarrow \{ \bm{\theta}_i^{(T_s)} | \bm{M}_i \small\text{ = }1\}$, \quad $\mathcal{F}_p \leftarrow \mathcal{F}(x|\hat{\bm{\theta}}, \bm{\zeta}^{(T_s)}) $ 
    
    \textcolor{gray}{\texttt{\# Further finetune the pruned subnet $\mathcal{F}(x|\hat{\bm{\theta}}, \bm{\zeta}^{(T_s)})$ with the original optimizer}}
    
    \For{$t\gets0$ \KwTo $T_{r} - 1$} {
        $ \hat{\bm{\theta}}^{(t+1)} \leftarrow \hat{\bm{\theta}}^{(t)} - \beta \frac{1}{n} \sum_{i=1}^{n} \nabla_{\hat{\bm{\theta}}} \mathcal{L_{O}}(\hat{\bm{\theta}}^{(t)})$ 
    }
    
    \Return $ \mathcal{F^{\star}} \leftarrow \mathcal{F}_p(x|\hat{\bm{\theta}}^{(T_r)})$
    
\end{algorithm}
\end{minipage}
\end{figure}

% \newpage
\subsection{Implentation of Unified Pruning}
\label{IMPLEMENTATION OF UNIFIED PRUNING}

The \textit{Unified Pruning} is outlined in Algorithm \ref{algorithm unified}. Line 1 $\sim$ 11 implements the search phase, and Line 12 $\sim$ 14 implements the retrain phase.

\begin{figure}[ht]
  \centering
  \vspace{-10pt}
\begin{minipage}[b]{\linewidth}
\begin{algorithm}[H]
    % \small
    \caption{Unified Pruning}
    \label{algorithm unified}
    \setcounter{AlgoLine}{0}
    \LinesNumbered

    \KwIn{Original model $\mathcal{F}$, parameters of the original model $\bm{\theta}$, parameters of the trainable mask $\bm{\zeta}$, total compression ratio $p$, iterations in the search stage $T_{s}$ and retrain stage $T_{r}$, learning rate for the search stage $\alpha$ and retrain stage $\beta$ }

    \KwOut{Model $\mathcal{F^{\star}}$ after the search and retrain}
    
    \For{$t\gets0$ \KwTo $T_{s} - 1$} {

        \textcolor{gray}{\texttt{\# Calculate the loss $\mathcal{L}$}}
        
        $\mathcal{L} \leftarrow \mathcal{L_{O}} + w_{a} \sum_{\bm{\zeta}_i \in \bm{\zeta}_a} \lVert \bm{\zeta}_i \rVert_{1} + w_{m} \sum_{\bm{\zeta}_i \in \bm{\zeta}_m} \lVert \bm{\zeta}_i \rVert_{1}$ 

        \textcolor{gray}{\texttt{\# Normally update both $\bm{\theta}$ and $\bm{\zeta}$ with the original optimizer}}
        
        $ \bm{\theta}^{(t+1)} \leftarrow \bm{\theta}^{(t)} - \alpha \frac{1}{n} \sum_{i=1}^{n} \nabla_{\bm{\theta}} \mathcal{L}(\bm{\theta}^{(t)}, \bm{\zeta}^{(t)})$, \quad 
        $ \bm{\zeta}^{(t+1)} \leftarrow \bm{\zeta}^{(t)} - \alpha \frac{1}{n} \sum_{i=1}^{n} \nabla_{\bm{\zeta}} \mathcal{L}(\bm{\theta}^{(t)}, \bm{\zeta}^{(t)})$ 

    }
    
    \textcolor{gray}{\texttt{\# Conduct z-score standardization to make $\bm{\zeta}_a$ and $\bm{\zeta}_m$ after the search comparable}}
    
    $\bm{\zeta}^{(T_s)}_a \leftarrow (\bm{\zeta}^{(T_s)}_a - \mathbb{E}[\bm{\zeta}^{(T_s)}_a]) / (\mathbb{E}[[\bm{\zeta}^{(T_s)}_a - \mathbb{E}[\bm{\zeta}^{(T_s)}_a]]^2])^{\frac{1}{2}}$, \quad 
    $\bm{\zeta}^{(T_s)}_m \leftarrow (\bm{\zeta}^{(T_s)}_m - \mathbb{E}[\bm{\zeta}^{(T_s)}_m]) / (\mathbb{E}[[\bm{\zeta}^{(T_s)}_m - \mathbb{E}[\bm{\zeta}^{(T_s)}_m]]^2])^{\frac{1}{2}}$

    \textcolor{gray}{\texttt{\# Generate pruning mask $\bm{M}$ by ranking and selecting on $\bm{\zeta}$}}
    
    $\bm{M} \leftarrow \small {\tt TopKMask}(\bm{\zeta}^{(T_s)}, \ p \cdot \small {\tt Size}(\bm{\zeta}))$

    \textcolor{gray}{\texttt{\# Compress $\bm{\theta}$ based on $M$ and accordingly compress $\mathcal{F}$}}
    
    $ \hat{\bm{\theta}} \leftarrow \{ \bm{\theta}^{(T_s)} | \bm{M} \small\text{ = }1\}$, \quad $\mathcal{F}_p \leftarrow \mathcal{F}(x|\hat{\bm{\theta}}, \bm{\zeta}^{(T_s)}) $ 

    \textcolor{gray}{\texttt{\# Further finetune the pruned subnet $\mathcal{F}(x|\hat{\bm{\theta}}, \bm{\zeta}^{(T_s)})$ with the original optimizer}}
    
    \For{$t\gets0$ \KwTo $T_{r} - 1$} {
        $ \hat{\bm{\theta}}^{(t+1)} \leftarrow \hat{\bm{\theta}}^{(t)} - \beta \frac{1}{n} \sum_{i=1}^{n} \nabla_{\hat{\bm{\theta}}} \mathcal{L_{O}}(\hat{\bm{\theta}}^{(t)})$ 
    }

    \Return $ \mathcal{F^{\star}} \leftarrow \mathcal{F}_p(x|\hat{\bm{\theta}}^{(T_r)})$

\end{algorithm}
\end{minipage}
\end{figure}

\section{Supplementary Related Works}
\label{more related work}

\textbf{Global pruning} Movement Pruning \cite{sanh2020movement} prunes unimodal BERT \cite{devlin2018bert} and investigates the global pruning, which conducts ranking on the whole model. The differences between it and UPop are: 1) It only prunes across different structures in unimodality language, while UPop not only prunes across different structures but also prunes across modalities. Therefore, it and UPop are complementary in terms of scope. 2) It finds local (ranking inside each weight matrix) and global pruning perform similarly on language tasks. However, UPop finds separate and unified pruning performs notably differently on multimodal tasks. And the different conclusions may be attributed to the intrinsic properties of different modalities and could be a meaningful topic for future research. 3) It treats all weights equally when ranks mask of different structures (\ie, Self-Attentions and MLPs). However, UPop notices that simply unified search space of different structures fail and proposes a solution based on z-score standardization to tackle this problem.

\textbf{Iterative pruning} Iterative pruning is an existing technique used by some prior works \cite{sanh2020movement, liu2022win}. The differences between it and UPop are: 1) It divides weights into multiple groups, and each time only binarizes a smaller number of non-zero weights to 0. For example,  to achieve a 50$\%$ total compression ratio, it divides weights into five groups, and each time only binarizes 10$\%$ non-zero weights to 0. On the other hand, UPop prunes each weight to make them from the original value progressively converge to zero. For example, suppose there is a single weight with an original value of 2.8. Then UPop will progressively prune this value to 0 (\eg, 2.8 to 2.7 to 2.6 to $\cdot\cdot\cdot$ to 0.1 to 0) while iterative pruning will directly prune it from 2.8 to 0. 2) Its binary masks take value from the set $\{0, 1\}$ while real-value masks of UPop take value from the interval $[0, 1]$. 3) Iterative pruning in \cite{sanh2020movement, liu2022win} has to use the straight-through estimator \cite{bengio2013estimating} to approximate gradients of binary masks to ensure weights can be normally updated, while this is not needed for real-value masks used by UPop.

\textbf{Parameter-efficient Tuning} Parameter-efficient tuning includes but not limited to low-rank adaptation \cite{hu2021lora, hyeon2021fedpara}, prompt tuning \cite{liu2022p, liu2021p}, parameter sharing \cite{lan2019albert, shi2021multi}, adapters \cite{he2021effectiveness, sung2022vl}, and dropout \cite{fan2019reducing, shi2022heuristic}. The difference between them and UPop is that they are used for reducing learnable parameters during tuning instead of inference. In contrast, UPop is used for reducing the parameters and FLOPs during inference instead of tuning. 

\newpage

\section{Supplementary Experiments and Analyses}
\label{more exp}

\subsection{Variation of Compressible Components and Layers}
\label{Variation of Compressible Components and layers}
\begin{figure}[h]
    \centering
    % \vspace{-7pt}
    \captionsetup{font={small}}
    \setlength{\abovecaptionskip}{0cm}
    \includegraphics[width=\textwidth]{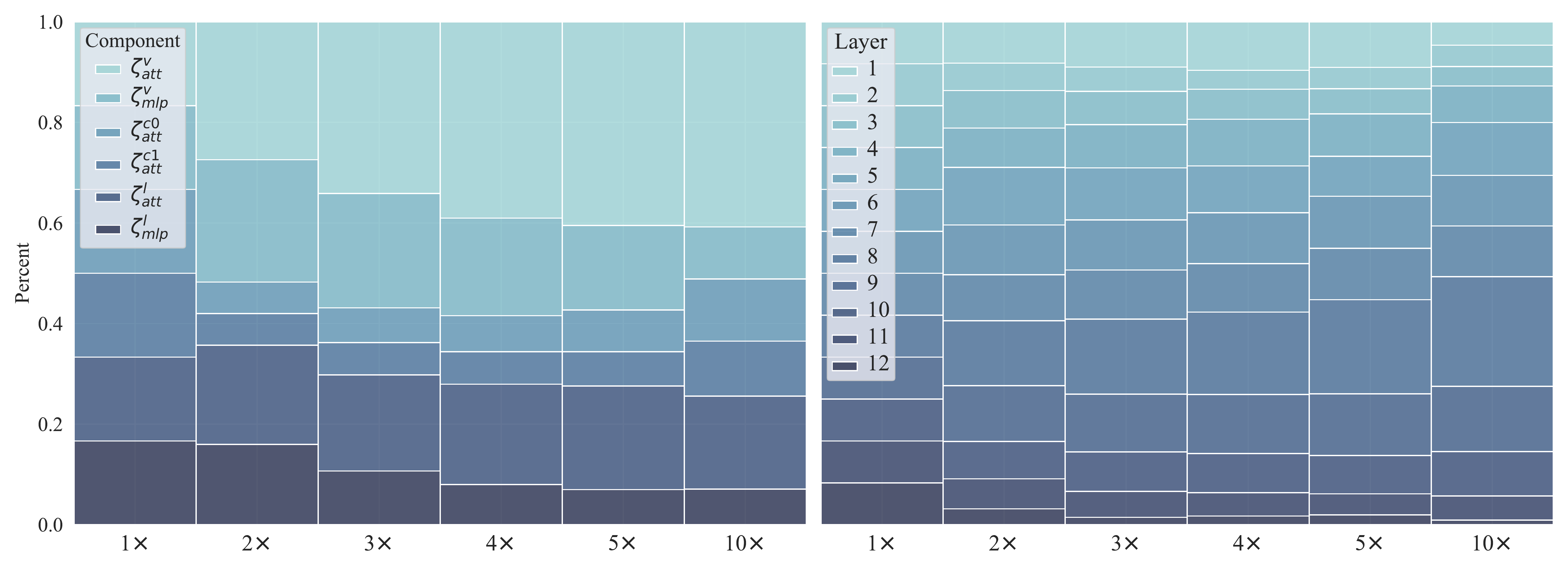}
    \caption{The left subfigure: variation of compressible components as the compression ratio increases. The right subfigure: variation of layers as the compression ratio increases.}
    \label{fig histogram}
    \vspace{-8pt}
\end{figure}

Unified Pruning enables the model to adaptively assign appropriate compression ratios among different compressible components. Accordingly, we demonstrate the variation of all components and layers as the total compression ratio increases in Figure \ref{fig histogram}. The left subfigure shows that the retained percentage of Self-Attention of ViT and Self-Attention of Bert among all compressible components significantly increases as the compression ratio increases. In contrast, the retained percentage of MLP of ViT and MLP of Bert decreases. This indicates that Self-Attentions have higher importance than MLPs when the number of parameters is limited. It can also be observed that vision modality is more important than language modality in this task. The trend of the retained percentage of Cross-Attention generally decreases and then increases. This phenomenon indicates that at low compression ratios, the parameters of the visual and language modalities are relatively adequate. Therefore cross-attention is less important at this time. At high compression ratios, the vision and language modality lacks sufficient parameters, and cross-attention becomes more critical. 

Similarly, the right subfigure of Figure \ref{fig histogram} demonstrates the variation of all layers as the total compression ratio increases. It can be observed that the middle layers occupy an increasing proportion as the total compression ratio increases, which indicates that the majority of modalities' information is generated in the middle layers of the model. In the earlier layers, the information is not detailed enough. In contrast, in the last several layers, the refinement of the information becomes less critical when the number of parameters is limited.

\subsection{Study on Update Strategy of Compression Ratio}
\label{update p_t appendix}
Compression ratio $p_t$ is a monotonically increasing function of iteration number $t$, and an intuitive design for updating $p_t$ is to increase $p_t$ evenly as $t$ increases, \ie:
\begin{equation}
\setlength{\abovedisplayskip}{0pt}
\setlength{\belowdisplayskip}{0pt}
\small
\label{equation p_t}
    p_t = p \frac{t}{T_s-1}
\end{equation}

It is worth noting that according to the implementation of Algorithm \ref{algorithm Upop}, the current compression ratio $p_t$ of $t^{th}$ iteration means that $p_t\%$ of embeddings has been compressed by $\frac{p_t}{p}\%$. As a consequence, the actual compression ratio $a_t$ should be the ratio of the compressed embedding size multiplied by the ratio of each embedding that is compressed:
\begin{equation}
\setlength{\abovedisplayskip}{3pt}
\setlength{\belowdisplayskip}{3pt}
\small
    a_t = p_t \times \frac{p_t}{p} = \frac{p_t^2}{p}
\end{equation}

In addition to the monotonically increasing property, a more appropriate update strategy than a uniform update strategy also needs to satisfy:
\vspace{-10pt}
\begin{itemize}
    \item On the one hand, the actual compression ratio should increase relatively slowly at the beginning of searching. Because when the iteration number $t$ is small, the cumulative gradients are relatively volatile, and the generated mask is relatively inaccurate.
    \item  On the other hand, the actual compression ratio should also increase relatively slowly toward the end of searching. Because as the current compression ratio gradually increases, the difficulty of compression also increases.
\end{itemize}

Formally speaking, $a_t$ is supposed to satisfy:
\begin{equation}
\setlength{\abovedisplayskip}{4pt}
\setlength{\belowdisplayskip}{4pt}
\label{equation requirements}
    \begin{cases}
    a_0 = 0\\
    a_{T_s-1} = p \\
	\frac{\mathrm{d} a_t}{\mathrm{d} t} \geq 0, \   \forall t \in [0, T_s-1] \\
	\exists\ t_0  \in (0, T_s) \ \ s.t. \ \frac{\mathrm{d}^2 a_t}{\mathrm{d} t^2} > 0,\ \forall t \in (0, t_0), \text{and} \ \frac{\mathrm{d}^2 a_t}{\mathrm{d} t^2} < 0,\ \forall t \in (t_0, T_s-1) \ \ 
	\end{cases}
\end{equation}
For example, the integration of trigonometric function $f(x)=\sin{\frac{\pi x}{T_s - 1}}$ defined on interval $[0, T_s -1]$ satisfies the latter two requirements of the Equation \ref{equation requirements}. To further satisfy the first two properties, we only need to let
\begin{equation}
    p \frac{\int_{0}^{t} \sin{\frac{\pi x}{T_s - 1}} \mathrm{d} x}{\int_{0}^{T_s - 1} \sin{\frac{\pi x}{T_s - 1}} \mathrm{d} x} = \frac{p}{2}(1-\cos{\frac{\pi t}{T_s - 1}}) = a_t = \frac{p_t^2}{p}
\end{equation}
And thus
\begin{equation}
    p_t = p(\frac{1}{2}(1-\small\text{cos}(\frac{\pi t}{T_s-1})))^{\frac{1}{2}}
\end{equation}
is a function that satisifies all requirements.

\begin{table}[h]
  \captionsetup{font={small}}
  \setlength{\abovecaptionskip}{0.2cm}
  \small
  \centering
  \caption{Study on how the update strategy of compression ratio $p_t$ affects the performance. The last one is adopted as our update strategy.}
  \begin{tabular}{l @{\hspace{1.0\tabcolsep}} | c @{\hspace{1.0\tabcolsep}} c}
  \toprule
    $p_t$ & Dev Acc & Test Acc \\
    \midrule
    \begin{small} $p \frac{t}{T_s-1}$ \end{small} & 79.94 & 80.84 \\
    \midrule
    \begin{footnotesize} $p \frac{(2T_s-t+1)t}{((T_s+1)T_s)} $ \end{footnotesize} & \textbf{80.38} & \textbf{81.13} \\
    \midrule
    \begin{small} $p (\frac{1}{2}(1-\small\text{cos}(\frac{\pi t}{T_s-1})))^{\frac{1}{2}} $ \end{small} &  80.33 &  \textbf{81.13} \\
    \bottomrule
  \end{tabular}
  % \vspace{-20pt}
  \label{table p_t}
\end{table}

Table \ref{table p_t} shows the performance of the 2 $\times$ compressed BLIP-NLVR model with different $p_t$ update strategies. The first one is the uniform update, while the last one is the strategy we adopted. There is obvious performance improvement when replacing the uniform update with $p (\frac{1}{2}(1-\small\text{cos}(\frac{\pi t}{T_s-1})))^{\frac{1}{2}}$. Besides, the last one is not the only feasible strategy, and other update strategies that satisfy requirements in Equation \ref{equation requirements} should also achieve better performance than uniform update. For example, the second strategy $p \frac{(2T_s-t+1)t}{((T_s+1)T_s)} $ also satisfies requirements and also achieves comparable performance to the strategy we adopted.

\subsection{Study on the frequency of Updating Compression Mask $\bm{\zeta}$}
We also explore how the frequency of updating mask $\bm{\zeta}$ affects the model performance. Experimental results on the $2\times$ compressed BLIP-NLVR model are reported in Table \ref{table interval}. Update compression mask $\bm{\zeta}$ at intervals has two benefits:
\begin{itemize}
    \item On the one hand, it can reduce a small amount of computation during searching.
    \item On the other hand, it can be observed from Table \ref{table interval} that updating the $\bm{\zeta}$ too frequently causes the compressed model to tend to overfit on the validation set.
\end{itemize}

\begin{table*}[h]
  \captionsetup{font={small}}
  \setlength{\abovecaptionskip}{0.2cm}
  \small
  \centering
  \caption{Study on how the frequency of updating compression mask $\bm{\zeta}$ affects the model performance. Frequency 50 is adopted by us.}
     \begin{tabular}{c|cc}
    \toprule
    Frequency & Dev Acc & Test Acc \\
    \midrule
    1 & \textbf{80.97} & 80.14 \\
    \midrule
    10 & 80.48 & 80.86 \\
    50 & 80.33 & \textbf{81.13} \\
    \bottomrule
  \end{tabular}
%   \vspace{-7pt}
  \label{table interval}
\end{table*}

The frequency $1$ means updating $\bm{\zeta}$ each time the model parameters $\bm{\theta}$ are updated, while frequency $10$ means updating $\bm{\zeta}$ once every 10 times the model parameters $\bm{\theta}$ are updated. Consequently, frequency 50 is adopted by us for the $2\times$ compressed BLIP-NLVR model, which mitigates the overfitting in the validation set and improves the performance on the test set. It is worth noting that the appropriate frequency varies for different models and tasks. Empirically, setting the frequency to the number of iterations corresponding to the $1\%$ compression ratio is more likely to be appropriate. For example, if we aim to accomplish $50\%$ compression ratio in 1000 iterations, then a frequency about $1000 \times \frac{1}{50} = 20$ should be recommended.

\subsection{Experiments on the Image Classification Task}
\label{exp classification appendix}

\begin{table*}[h]
  \setlength{\tabcolsep}{5 pt}
  \captionsetup{font={small}}
  \setlength{\abovecaptionskip}{0.2cm}
  \small
  \centering
  \caption{Compress DeiT on the ImageNet dataset. The units of Params and FLOPs are M and G, respectively. The superscript \textsuperscript{$*$} indicates the performance of the deployable model if the original model is non-deployable. For fairness of comparison, all reported experimental results, including UPop, do not use knowledge distillation.}
  \begin{tabular}{l|l @{\hspace{0.7\tabcolsep}} l|l @{\hspace{1.0\tabcolsep}} l}
    \toprule
    Approach & Top-1 (\%) & Top-5 (\%) & Params & FLOPs \\
    \midrule
    DeiT \cite{touvron2021training} & 79.9 & 95.0 & 22.0 & 4.6  \\
    GLiT \cite{chen2021glit} & 80.5 & - & 24.6 & 4.4 \\
    DynamicViT \cite{rao2021dynamicvit}& 79.3 & - & 22.0 & 2.9  \\
    S$^2$ViTE \cite{chen2021chasing} & 79.2 & - & 14.6 & 3.1 \\
    ViTAS \cite{su2022vitas} & 80.2 & 95.1 & 23.0 & 4.9  \\
    ViT-Slimming \cite{chavan2022vision} & 77.9  & 94.1 & 11.4 & 2.3  \\
    ViT-Slimming\textsuperscript{$*$} \cite{chavan2022vision} & 77.1 & 93.6 & 11.4 & 2.3  \\
    EViT \cite{liang2022not} & 78.5 & 94.2 & 22.0 & 2.3 \\
    A-ViT \cite{yin2022vit} & 78.6 & - & 22.0 & 3.6 \\
    \midrule
    DeiT with UPop$_{1.11\times}$ (Ours) & 81.1$_{\color{red}\uparrow 1.2}$ & 95.4$_{\color{red}\uparrow 0.4}$ & 19.9$_{\color{red}\downarrow 10\%}$ & 4.1$_{\color{red}\downarrow 11\%}$ \\
    DeiT with UPop$_{1.25\times}$ (Ours) & 80.8$_{\color{red}\uparrow 0.9}$ & 95.4$_{\color{red}\uparrow 0.4}$ & 17.8$_{\color{red}\downarrow 19\%}$ & 3.7$_{\color{red}\downarrow 20\%}$ \\
    DeiT with UPop$_{1.42\times}$ (Ours) & 80.2$_{\color{red}\uparrow 0.3}$ & 95.1$_{\color{red}\uparrow 0.1}$ & 15.7$_{\color{red}\downarrow 29\%}$ & 3.2$_{\color{red}\downarrow 30\%}$ \\
    DeiT with UPop$_{1.67\times}$ (Ours) & 79.6$_{\color{red}\downarrow 0.3}$ & 94.8$_{\color{red}\downarrow 0.2}$ & 13.5$_{\color{red}\downarrow 39\%}$ & 2.8$_{\color{red}\downarrow 39\%}$ \\
    DeiT with UPop$_{2.00\times}$ (Ours) & 78.9$_{\color{red}\downarrow 1.0}$ & 94.6$_{\color{red}\downarrow 0.4}$ & 11.4$_{\color{red}\downarrow 48\%}$ & 2.3$_{\color{red}\downarrow 50\%}$ \\
    \bottomrule
  \end{tabular}
  \label{table imagenet}
%   \vspace{-7pt}
\end{table*}

\subsection{Experiments on the Image Segmentation Task}
\label{exp classification appendix}

\begin{table*}[h]
  \setlength{\tabcolsep}{5 pt}
  \captionsetup{font={small}}
  \setlength{\abovecaptionskip}{0.2cm}
  \small
  \centering
  \caption{Compress Segmenter on the ADE20k dataset. The units of Params and FLOPs are M and G, respectively. The SS and MS mean single-scale and multi-scale testing for the mIoU metric, respectively. With and Without superscript \textsuperscript{$*$} means CNN-based and Transformer-based models, respectively.}
  \begin{tabular}{l|l @{\hspace{0.7\tabcolsep}} l|l @{\hspace{1.0\tabcolsep}} l}
    \toprule
    Approach & mIoU (SS) & mIoU (MS) & Params & FLOPs \\
    \midrule
    Segmenter \cite{strudel2021segmenter} & 45.3 & 46.9 & 26.4 & 38.6  \\
    Swin Transformer \cite{liu2021swin} & 44.5 & 46.1 & 60.0 & 236.0 \\
    SenFormer \cite{bousselham2021efficient} & - & 46.0 & 59.0 & 179.0 \\
    SegFormer \cite{xie2021segformer} & 46.5 & 47.5 & 27.5 & 62.4 \\
    PVT \cite{wang2021pyramid} & 39.8 & - & 28.2 & 44.5 \\
    PVTv2 \cite{wang2022pvt} & 45.2 & - & 29.1 & 45.8 \\
    DeiT III \cite{touvron2022deit} & 45.6 & 46.8 & 41.7 & - \\
    DeeplabV3+ (ResNet-101)\textsuperscript{$*$} \cite{chen2018encoder} & 45.5 & 46.4 & 63.0 & 255.0 \\
    OCRNet (HRNet-W48)\textsuperscript{$*$} \cite{yuan2020object} & - & 45.7 & 70.5 & 164.8 \\
    ConvNeXt \textsuperscript{$*$} \cite{liu2022convnet} & - & 46.7 & 60.0 & - \\
    Segmenter with Mask-based Pruning \cite{chavan2022vision} & 38.7 & 41.0 & 18.9 & 27.3 \\
    \midrule
    Segmenter with UPop$_{1.10\times}$ (Ours) & 45.9$_{\color{red}\uparrow 0.6}$ & 47.6$_{\color{red}\uparrow 0.7}$ & 23.9$_{\color{red}\downarrow 10\%}$ & 33.7$_{\color{red}\downarrow 13\%}$ \\
    Segmenter with UPop$_{1.16\times}$ (Ours) & 45.5$_{\color{red}\uparrow 0.6}$ & 47.3$_{\color{red}\uparrow 0.4}$ & 22.7$_{\color{red}\downarrow 14\%}$ & 32.0$_{\color{red}\downarrow 17\%}$ \\
    Segmenter with UPop$_{1.23\times}$ (Ours) & 45.3$_{\color{red}\uparrow 0.0}$ & 47.1$_{\color{red}\uparrow 0.2}$ & 21.5$_{\color{red}\downarrow 19\%}$ & 30.4$_{\color{red}\downarrow 21\%}$ \\
    Segmenter with UPop$_{1.39\times}$ (Ours) & 44.4$_{\color{red}\downarrow 0.9}$ & 46.3$_{\color{red}\downarrow 0.6}$ & 18.9$_{\color{red}\downarrow 28\%}$ & 27.6$_{\color{red}\downarrow 29\%}$ \\
    \bottomrule
  \end{tabular}
  \label{table ade20k}
%   \vspace{-7pt}
\end{table*}

We also conduct experiments on the unimodal Segmenter-S model and ADE20k dataset for the semantic segmentation task as shown in Table \ref{table ade20k}, where we compare Segmenter pruned by UPop with the uncompressed Segmenter and other models with similar params/FLOPs to our pruned models. Furthermore, we also compare the performance of UPop on different unimodal tasks. Experimental results demonstrate:
% \vspace{-7pt}
\begin{itemize}
    \item When comparing the model pruned by UPop with the uncompressed model, UPop can achieve more than 1.2$\times$ loss-free compression and around 1.4$\times$ compression with less than 1$\%$ mIoU loss for both single-scale and multi-scale testing.
    \item When comparing the model pruned by UPop with other models with similar params/FLOPs, UPop can achieve very competitive performance under Performance-Parameters and Performance-FLOPs trade-off constraints. For example, the 1.1$\times$ compressed model can outperform all other models on multi-scale testing, and achieve a second place on single-scale testing by only using 54$\%$ FLOPs of the first place model.
    \item When comparing UPop's performance on the semantic segmentation task with on the image classification task (refers to Appendix Table \ref{table imagenet}), UPop can achieve around 1.5$\times$ loss-free compression and 2$\times$ compression with no more than 1$\%$ accuracy loss on the image classification task. We believe this ratio gap should be attributed to the intrinsic properties of different tasks. For example, classification models have more redundancy since many pixels (\eg, background pixels) are unimportant for classification results, and therefore classification models can be pruned more easily. On the other hand, segmentation models have less redundancy since the model is expected to output the corresponding category for each pixel, and therefore pruning segmentation models is more difficult.
\end{itemize}

\end{document}